\PassOptionsToPackage{table}{xcolor}
\documentclass[10pt,a4paper,logo]{googledeepmind}
\usepackage{times}
\usepackage{natbib}
\setcitestyle{authoryear,round,citesep={;},aysep={,},yysep={;}}
\newcommand{\paperCodeURL}{https://github.com/Eureka-Maggie/SyncPro}
\newcommand{\casepanelwidth}{0.88\linewidth}

\usepackage{amsmath,amsfonts,bm}

\def\eqref#1{equation~\ref{#1}}

\def\1{\bm{1}}

\DeclareMathAlphabet{\mathsfit}{\encodingdefault}{\sfdefault}{m}{sl}
\SetMathAlphabet{\mathsfit}{bold}{\encodingdefault}{\sfdefault}{bx}{n}

\usepackage{hyperref}
\usepackage{url}
\usepackage{booktabs}
\usepackage{multirow}
\usepackage[table]{xcolor}
\usepackage{graphicx}
\usepackage{subcaption}
\usepackage{colortbl}
\usepackage{float}
\usepackage{placeins}
\usepackage[most]{tcolorbox}
\usepackage{listings}
\usepackage{xspace}
\newcommand{\ours}{\textsc{SyncPro}\xspace}
\newcommand{\innername}{\textit{Inner}\xspace}
\newcommand{\responsename}{\textit{Response}\xspace}
\newcommand{\oursrl}{\textsc{SyncPro-RL}\xspace}
\newcommand{\ourssft}{\textsc{SyncPro-SFT}\xspace}
\newcommand{\interleaved}{\textit{Interleaved}\xspace}
\newcommand{\responseonly}{\textit{Response-only}\xspace}
\newcommand{\best}[1]{\textbf{#1}}
\newcommand{\second}[1]{\underline{#1}}

\definecolor{tableblue}{HTML}{EAF4F8}
\definecolor{tablepurple}{HTML}{F2EDF8}
\definecolor{tablepeach}{HTML}{FBEFE7}

\newtcblisting{promptbox}[1]{
  enhanced,
  listing engine=listings,
  listing only,
  title={#1},
  colback=gray!6,
  colframe=gray!35,
  boxrule=0.4pt,
  arc=1mm,
  left=1mm,
  right=1mm,
  top=0.8mm,
  bottom=0.8mm,
  before skip=5pt,
  after skip=7pt,
  fonttitle=\bfseries\small,
  listing options={
    basicstyle=\ttfamily\scriptsize,
    breaklines=true,
    columns=fullflexible,
    keepspaces=true,
    showstringspaces=false
  }
}

\title{SyncPro: Synchronous Reasoning and Responding for Proactive Streaming Video Understanding}
\renewcommand{\titlefont}{\color{black}\normalfont\bfseries\fontsize{19}{22}\selectfont}

\author[1,2,3]{Xueyun Tian}
\author[1]{Bingbing Xu}
\author[3]{Jiaming Liu\textsuperscript{\textdagger}}
\author[3]{Jinpeng Yu}
\author[3]{Yang Shi}
\author[1]{Yu Zhao}
\author[3]{Ruihua Huang}
\author[1]{Yuanzhuo Wang}
\author[1,2]{Huawei Shen}
\affil[1]{State Key Laboratory of AI Safety, Institute of Computing Technology, CAS}
\affil[2]{University of Chinese Academy of Sciences}
\makeatletter
\def\AB@affilsep{\protect\\[1pt]}
\makeatother
\affil[3]{Qwen Business Unit of Alibaba}
\definecolor{SyncProlink}{HTML}{28548A}
\newcommand{\SyncProresourcelink}[4]{%
  \href{#1}{\mbox{%
    \makebox[1.4em][c]{\raisebox{-.2\height}{\includegraphics[height=#3]{#2}}}%
    \hspace{.25em}\textcolor{SyncProlink}{\textbf{#4}}}}}
\footnotetext{\textsuperscript{\textdagger}\,Project Leader.\hspace{1.5em}%
Email: \href{mailto:tianxueyun23z@ict.ac.cn}{\textcolor{black!70}{tianxueyun23z@ict.ac.cn}}\\[6pt]
{\fontsize{9.5}{12}\selectfont
\SyncProresourcelink{https://huggingface.co/datasets/EurekaTian/SyncPro-189k}{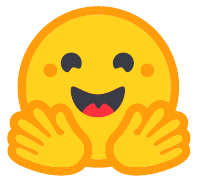}{1.14em}{Data}\hspace{1.8em}%
\SyncProresourcelink{https://huggingface.co/EurekaTian/SyncPro-RL}{figures/huggingface-logo.pdf}{1.14em}{Model}\,(\href{https://huggingface.co/EurekaTian/SyncPro-SFT}{\textcolor{SyncProlink}{SFT}} / \href{https://huggingface.co/EurekaTian/SyncPro-RL}{\textcolor{SyncProlink}{RL}})\hspace{1.8em}%
\SyncProresourcelink{https://github.com/Eureka-Maggie/SyncPro}{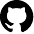}{1.08em}{Code}\hspace{1.8em}%
\SyncProresourcelink{https://eureka-maggie.github.io/SyncPro/}{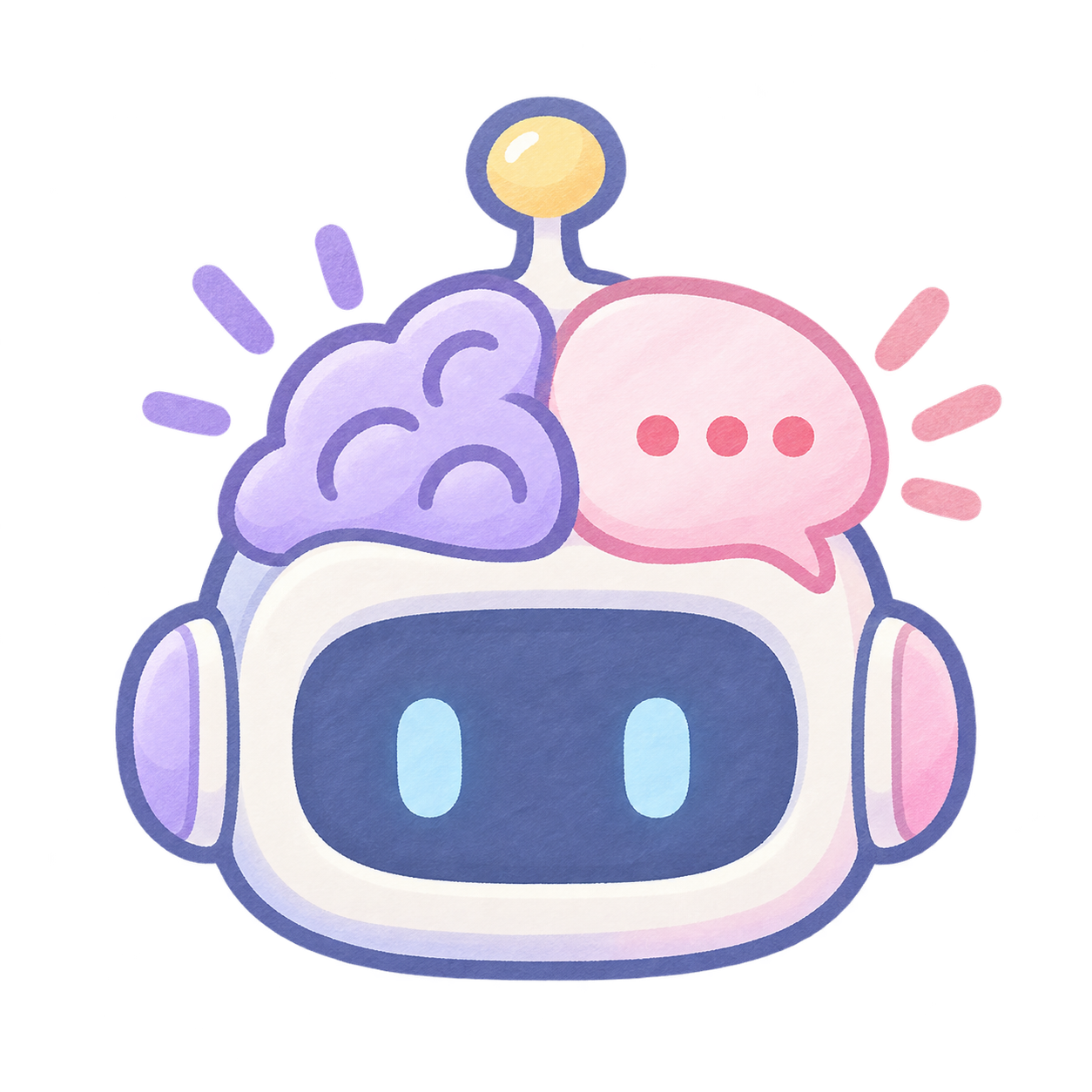}{1.5em}{Project Page}}}

\begin{abstract}
Proactive streaming video understanding requires assistants to track unfolding events and respond when a user's requested condition is met.
To make these decisions, assistants must accumulate relevant evidence and reassess response conditions as new observations arrive.
Existing work offers complementary capabilities for this process: proactive video assistants learn when to respond, while streaming video reasoning methods generate scheduled intermediate text to organize observations and support subsequent answers.
Together, these capabilities provide a foundation for proactive reasoning, but how to coordinate ongoing reasoning with timely responses remains underexplored.
In a direct adaptation, fixed reasoning intervals can trigger updates when nothing relevant has changed or postpone useful updates until the next scheduled step.
Sharing one generation path also couples two distinct requirements: preserving evidence for future decisions and producing only the response requested by the user.
Moreover, responses generated after reasoning must wait for it to finish, even when sufficient evidence is already available.
To address these challenges, we introduce \ours{}, a synchronous dual-stream framework with specialized generation paths for evidence tracking and user-facing responses.
Its \innername{} and \responsename{} branches share causal history and activate independently, allowing relevant evidence changes to trigger internal updates and responses to proceed concurrently with ongoing reasoning.
We construct 189K dual-stream examples and train \ours{} through supervised fine-tuning followed by trajectory-level reinforcement learning.
Experiments across three proactive benchmarks show improved response timing and more accurate answers to user requests.
Code will be released at \href{\paperCodeURL}{this repository}.

\end{abstract}

\begin{document}

\maketitle

\section{Introduction}
\label{sec_intro}
Proactive video assistants respond as events unfold \citep{qian2025dispider,wang2025videollm,yan2026proactvl}, rather than after receiving complete videos \citep{maaz2024videochatgpt,lin2024videollava}.
This requires tracking query-relevant evidence and assessing response conditions.
Explicit reasoning can support this monitoring through internal updates that are not shown to the user, while responses deliver the requested information or assistance.
The challenge is to coordinate these activities while accommodating their distinct roles and timing.

Two lines of work provide complementary foundations: proactive assistants learn response timing over video streams \citep{chen2024videollmonline,qian2025dispider}, while streaming reasoning methods generate intermediate text as observations arrive \citep{guan2026video,zhang2026think}.
Recent work combines these capabilities: ThinkStream uses incremental reasoning to decide whether to continue observing or respond \citep{liu2026thinking}.
However, it follows a fixed schedule of reasoning at every input step, with reasoning and response decisions arranged serially in one generation sequence.

\begin{figure*}[t]
    \centering
    \includegraphics[width=\textwidth]{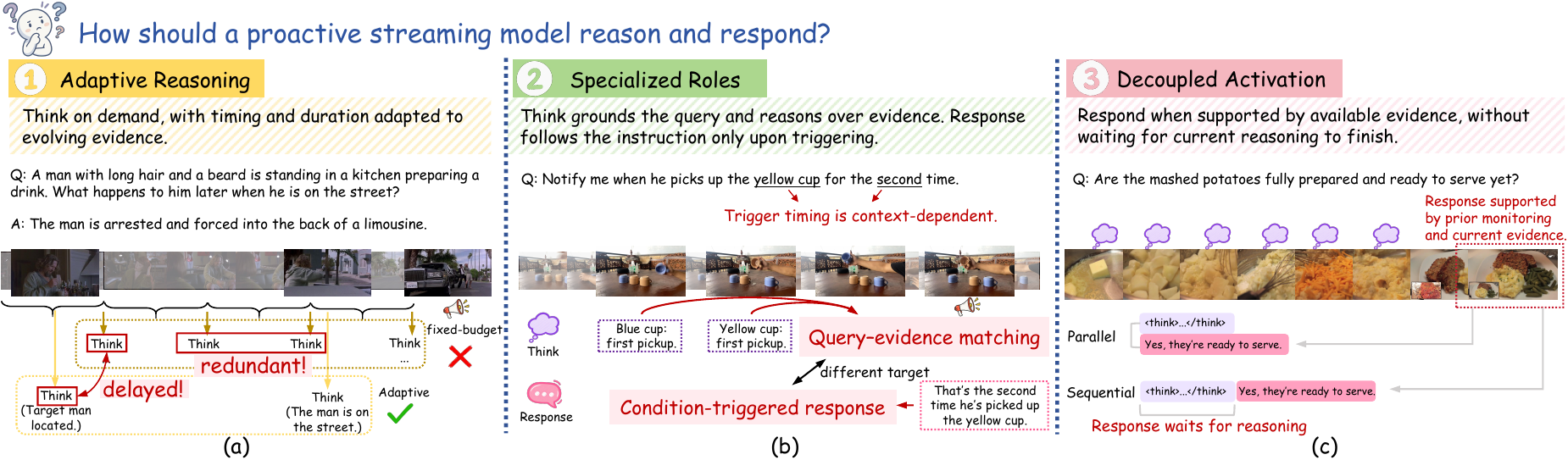}
    \caption{
        \textbf{Requirements for proactive streaming reasoning.}
        \textbf{(a) Adaptive reasoning:}
        internal updates follow query-relevant evidence rather than a fixed schedule.
        \textbf{(b) Specialized roles:}
        reasoning supports event monitoring, while responses deliver the requested output.
        \textbf{(c) Decoupled activation:}
        responses supported by prior monitoring and current evidence need not await further reasoning.
    }
    \label{fig:intro}
\end{figure*}

The central challenge is to coordinate ongoing reasoning with proactive responses as evidence evolves. This coordination involves three aspects.
First, reasoning should adapt its timing and duration to relevant evidence (Figure~\ref{fig:intro}(a)). Fixed schedules can cause redundant or delayed updates.
Second, reasoning should maintain query-relevant state, while responses should fulfill the user's request (Figure~\ref{fig:intro}(b)).
Finally, the roles need independent activation (Figure~\ref{fig:intro}(c)). Responses supported by available evidence should proceed alongside ongoing reasoning.

We therefore introduce \ours{}, a synchronous dual-stream framework on a shared multimodal backbone.
To support adaptive reasoning, each branch decides whether to generate or remain silent based on the shared history and current observation.
Separate low-rank adapters specialize \innername{} for evidence tracking and \responsename{} for user-facing responses.
Independent activation lets both branches decode concurrently without accessing each other's current outputs.
Their completed outputs enter the shared history for later decisions.
This design coordinates ongoing reasoning with proactive responses while preserving their distinct roles and timing.

We construct 189K dual-stream examples and train \ours{} through supervised fine-tuning and reinforcement learning.
On the Proactive Output task of StreamingBench \citep{lin2026streamingbench}, it achieves 41.2\% accuracy within a two-second response window, exceeding the strongest compared baseline by 3.2 percentage points.
It also outperforms prior methods on all three Forward Active Responding tasks of OVO-Bench \citep{niu2025ovo}, with an average F1 score of 34.19\%, and obtains the highest PAUC score of 32.43\% among the compared methods on the VAD subset of ProactiveVideoQA \citep{wang2025proactivevideoqa}.
Ablations and analyses validate explicit reasoning and role separation, with specialized branches and largely preserved offline video understanding.



Our main contributions are:
\begin{itemize}
    \item \textbf{Synchronous reasoning and responding.}
    We introduce a dual-stream framework for proactive streaming video understanding. Its specialized streams activate independently and jointly update a shared history for subsequent decisions.

    \item \textbf{Dual-stream data and training.}
    We provide 189K trajectories with per-step supervision for internal updates and user-facing responses. Supervised fine-tuning learns both behaviors, and trajectory-level reinforcement learning refines response timing.

    \item \textbf{Empirical evaluation and analysis.}
    Experiments on three proactive benchmarks demonstrate gains in response timing and content. Controlled comparisons support dual-stream modeling over response-only and interleaved generation.
\end{itemize}

\section{Related Work}
\label{sec_related_works}
\paragraph{Proactive Video Understanding.}
Offline video models respond after observing complete videos \citep{maaz2024videochatgpt,lin2024videollava}, whereas streaming models ground outputs in the observed video prefix \citep{chen2024videollmonline,huang2025onlinevideo}.
Proactive methods further regulate response timing through explicit control mechanisms or learned policies \citep{wang2025videollm,wang2026mmduet2,xueyun2026roma,yan2026proactvl,azad2026streamready,ma2026responseg1,yang2026livestarpro}.
Policy optimization incorporates class reweighting and rewards over turns and trajectories \citep{li2026streampro}, while sequential control coordinates memory compression with response readiness and adaptive reasoning \citep{liu2026r3streaming}.
Both formulations retain a single user-facing output stream.

\paragraph{Streaming Video Reasoning.}
Streaming reasoning methods incorporate intermediate reasoning and responses into a single autoregressive sequence as observations arrive \citep{zhao2026cogstream,guan2026video,zhang2026think,wang2026think}.
Accumulated reasoning traces also guide response decisions throughout the stream \citep{liu2026thinking}.
Such formulations often schedule reasoning at predefined processing intervals and generate responses after the associated reasoning segment.
This couples response generation to the timing and completion of reasoning.

\paragraph{Concurrent Multimodal Interaction.}
Existing work introduces concurrency at different levels.
At the output level, parallel generation of semantic and acoustic tokens produces a single spoken response \citep{xu2025qwen2,xu2025qwen3}.
Input--output duplex models extend this overlap to perception during response generation while retaining a single semantic output stream \citep{lin2026speak,cui2026minicpm}.
Beyond perception and generation, concurrency also supports internal coordination.
Reasoning and interaction modules communicate asynchronously \citep{huang2026duplexomni}, while text agents use parallel role-specific streams to solve a shared task \citep{su2026multistream}.
\ours{} instead separates ongoing event monitoring from condition-triggered task execution.
The two streams share causal history but activate independently without accessing each other's outputs at the current step.

\section{Method}
\label{sec_method}
\subsection{Task Formulation}
Let $q$ be a user instruction specifying a monitoring objective and the required response. A video arrives as a causal sequence $\mathcal{V}=(v_1,\ldots,v_T)$. Each observation $v_t$ defines micro-turn $t$. Under our 1-fps experimental protocol, each $v_t$ is one sampled video frame. The model receives $c_t=(q,\mathcal{H}_{t-1},v_t)$, where $\mathcal{H}_{t-1}$ contains information retained from previous micro-turns. Future observations are unavailable during generation.

We model internal reasoning and user-facing responses as two generation branches. \innername{} ($I$) tracks query-relevant evidence, while \responsename{} ($R$) produces the requested output. At each micro-turn, each branch chooses whether to remain silent or generate content:
\begin{equation}
a_t^{I}\in\mathcal{A}_{I}
=\{\langle\mathrm{silent}\rangle,\langle\mathrm{inner}\rangle\},
\qquad
a_t^{R}\in\mathcal{A}_{R}
=\{\langle\mathrm{silent}\rangle,\langle\mathrm{speak}\rangle\}.
\label{eq:branch_actions}
\end{equation}
Let $o_t^{I}$ and $o_t^{R}$ denote the complete output sequences of the two branches. Each branch attends only to $c_t$ and its own preceding tokens. The outputs follow branch-specific conditional distributions:
\begin{equation}
o_t^{I}\sim p_{I}(\cdot\mid c_t),
\qquad
o_t^{R}\sim p_{R}(\cdot\mid c_t),
\label{eq:parallel_decoding}
\end{equation}

Each output begins with an action token. A $\langle\mathrm{silent}\rangle$ action terminates the sequence immediately. A $\langle\mathrm{inner}\rangle$ or $\langle\mathrm{speak}\rangle$ action is followed by branch-specific autoregressive content. Both branches may activate together when the current observation requires state tracking and a user-facing response. Only active \responsename content is returned to the user during interaction.

The two streams synchronize at micro-turn boundaries. Once both finish, their complete outputs jointly update the shared context for the next observation:
\begin{equation}
\mathcal{H}_t
=
U(\mathcal{H}_{t-1},o_t^{I},o_t^{R}).
\label{eq:history_update}
\end{equation}
Here $U$ incorporates both output sequences and their layer-specific attention states. The visual tokens of $v_t$ are then discarded. Both outputs remain in the history and inform later decisions.

\begin{figure*}[t]
\centering
\includegraphics[width=1.0\textwidth]{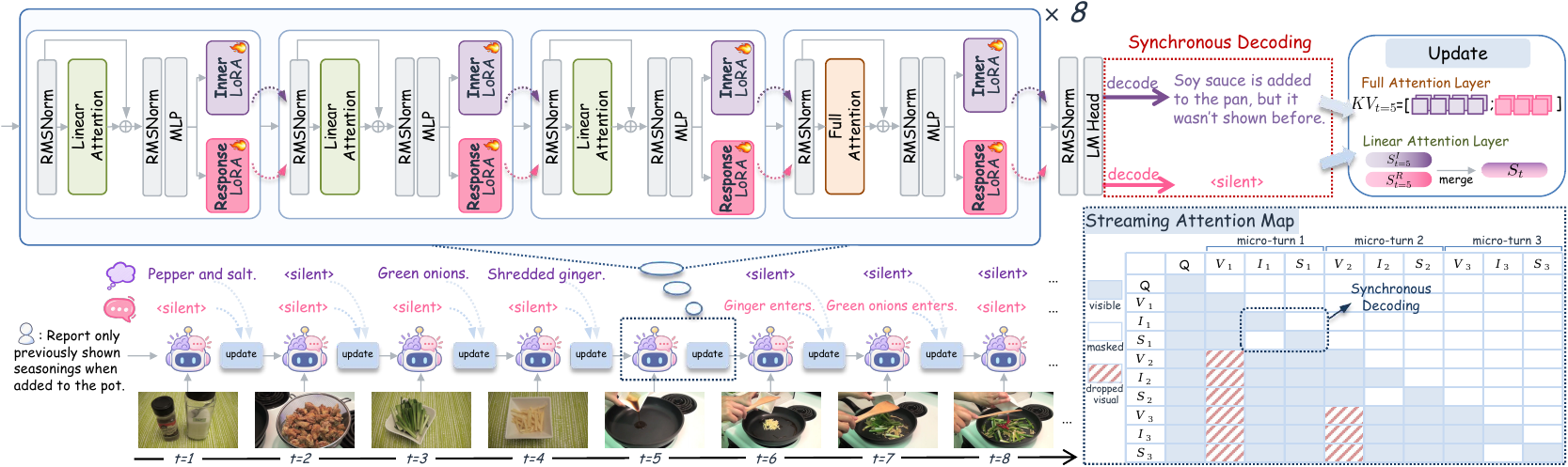}
\caption{\textbf{Dual-stream modeling in \ours.} At each micro-turn, \innername and \responsename select actions independently and decode from the same multimodal prefix through separate LoRA routes. Only \responsename is user-visible. Both outputs update the context for the next micro-turn before the visual tokens are discarded to limit persistent cache growth.}
\label{fig:model}
\end{figure*}

\subsection{\ours{} Model Architecture}
\label{sec:model_architecture}
We propose \ours{} to coordinate ongoing reasoning with timely responses. As shown in Figure~\ref{fig:model}, it combines a shared multimodal backbone with two specialized branches. \innername{} maintains evidence for later decisions, while \responsename{} delivers the requested output. Independent branch activation allows supported responses to proceed alongside ongoing reasoning.

At each micro-turn, the backbone encodes $c_t$ once for both branches. Separate low-rank adaptation (LoRA) modules in each decoder layer's MLP projections parameterize their action and content distributions \citep{hu2022lora}, implementing Equation~\ref{eq:parallel_decoding} with independent decoding states. After generation, \ours{} appends $o_t^{I}$ then $o_t^{R}$ to the shared history and merges their layer states.

For a full-attention layer, let $(K_t^{0},V_t^{0})$ denote the shared-prefix KV cache and $(K_t^{b},V_t^{b})$ the cache increment for branch $b$'s independently decoded action and content tokens. Concatenating these caches retains both outputs for the next micro-turn:
\begin{equation}
K_t=[K_t^{0};K_t^{I};K_t^{R}],
\qquad
V_t=[V_t^{0};V_t^{I};V_t^{R}],
\label{eq:full_attention_merge}
\end{equation}
where semicolons denote sequence concatenation. The block-causal mask enforces same-prefix decoding without same-turn cross-branch attention. Following prior position-sharing designs \citep{chen2025aspd,xueyun2026roma}, both increments use the first MRoPE position after the prefix to mark a common temporal origin. Appendix~\ref{app:implementation} details the mask and position rules.

Qwen3.5 inherits Qwen3-Next's hybrid decoder, whose linear-attention layers use Gated DeltaNet \citep{qwen2026qwen35,yang2025gated}. In each such layer, the shared prefix yields recurrent state $S_t^{0}$, and independent decoding yields $S_t^{b}$ for branch $b$. Appendix~\ref{app:implementation} gives the inherited recurrence. We define $\Delta S_t^{b}=S_t^{b}-S_t^{0}$ and merge both branch updates:
\begin{equation}
S_t
=
S_t^{0}
+\gamma\left(\Delta S_t^{I}+\Delta S_t^{R}\right),
\label{eq:linear_attention_merge}
\end{equation}
With $\gamma=0.5$, Equation~\ref{eq:linear_attention_merge} reduces to $S_t=(S_t^I+S_t^R)/2$. This symmetric heuristic assigns neither semantic role prior weight and is invariant to branch order. The transition generally differs from serial composition. Using the same rule during training and inference lets the LoRA routes learn subsequent token distributions under the merged recurrent states. The concatenated text and merged layer states together instantiate the history-update operator $U$ in Equation~\ref{eq:history_update}.

\paragraph{Streaming inference.}
\label{sec:streaming_inference}
We encode each frame once and fork the shared state for parallel branch decoding. Only active \responsename{} content is streamed to the user. We then merge both branch states and discard frame tokens before the next frame. Appendix~\ref{app:implementation} gives implementation details.

\FloatBarrier

\begin{figure*}[t]
    \centering
    \begin{subfigure}[c]{0.695\textwidth}
        \centering
        \includegraphics[width=\linewidth]{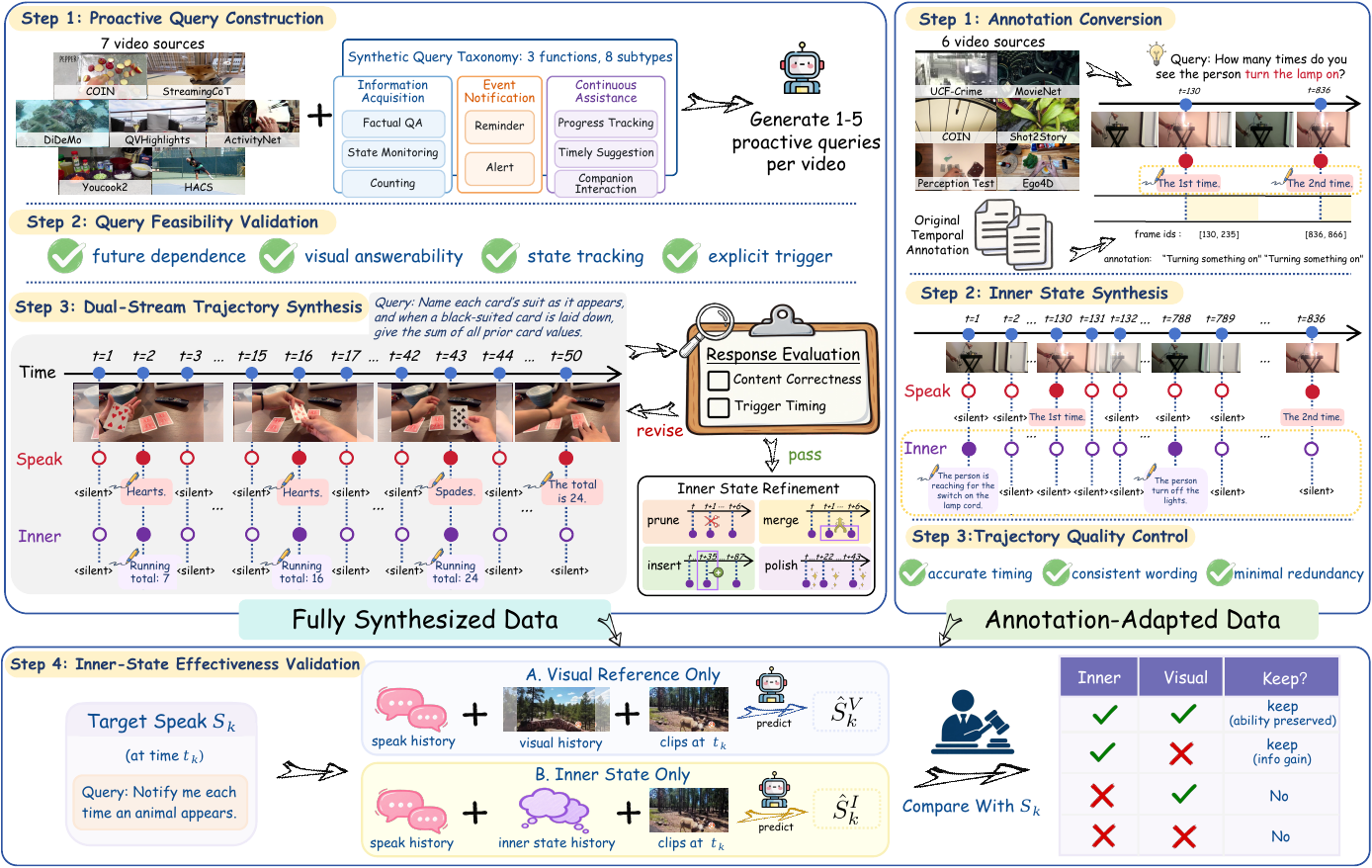}
        \caption{Construction pipeline.}
        \label{fig:pipeline}
    \end{subfigure}
    \hfill
    \begin{subfigure}[c]{0.295\textwidth}
        \centering
        \includegraphics[width=\linewidth]{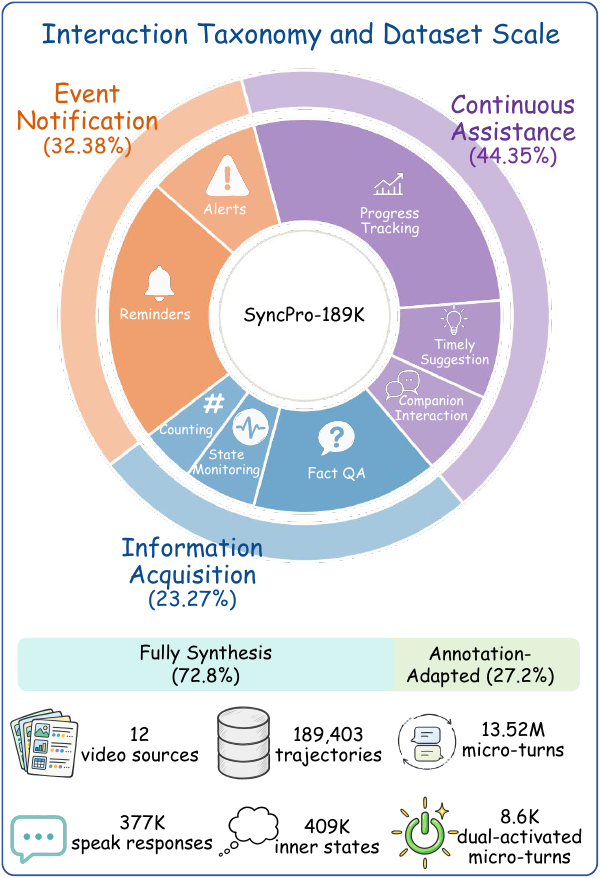}
        \caption{Corpus composition.}
        \label{fig:data}
    \end{subfigure}
    \caption{\textbf{Construction of \ours-189K.} Two routes produce causal dual-stream trajectories. A synthesized trajectory is retained only when its accumulated \innername entries suffice to reconstruct every later response from restricted visual context. The resulting 189,403 trajectories span three interaction functions, eight subtypes, and 12 video sources.}
    \label{fig:construct_data}
\end{figure*}
\subsection{Dual-Stream Data Construction}
\label{sec_data}
Each training example pairs a query and video with a dual-stream trajectory. We label each branch's action at every micro-turn and provide text targets for active branches. Existing datasets lack these paired targets \citep{hu2025streamingcot,guan2026video,liu2026thinking}. We construct these trajectories through proactive synthesis and temporal-label adaptation (Figure~\ref{fig:pipeline}).

The first route builds proactive queries from seven video sources \citep{krishna2017dense_captioning,lei2021qvhighlights,tang2019coin,zhou2018youcook2,zhao2019hacs,hendricks2017didemo,hu2025streamingcot}. We retain queries requiring future evidence, state tracking, and an observable trigger. Each target at $t$ is generated from the physically truncated prefix $\mathcal{V}_{\leq t}$, while other micro-turns receive silent actions. An offline evaluator audits the full video for content, timing, and event coverage. We allow up to three revisions under the corresponding prefix, then remove redundant entries, insert missing event-boundary states, and normalize language. Synthesized \innername{} and \responsename{} entries occupy distinct micro-turns, with the final \innername{} update preceding its \responsename{}. This avoids repeating trigger evidence already preserved by \responsename{} in shared history.

The second route maps temporal annotations from six datasets to this format \citep{tang2019coin,han2025shot2story,grauman2022ego4d,huang2020movienet,patraucean2023perception_test,sultani2018ucf_crime}. Event and answer intervals determine response timing and content. Query-conditioned generation inserts \innername at relevant state changes and preserves coincident boundaries when a state change matches response onset. We use only official training splits, with video IDs disjoint from the held-out ProactiveVideoQA test videos \citep{wang2025proactivevideoqa}.

We evaluate synthesized \innername{} updates by reconstructing responses under two contexts. The visual-reference context contains prior responses and all frames since the preceding response. The inner-state context contains prior \innername and \responsename entries but only the current visual window. An independent semantic judge evaluates both reconstructions. We retain trajectories only when the inner-state context reconstructs every response after earlier visual evidence is removed.

The corpus contains 189,403 trajectories from 12 sources across three functions and eight subtypes (Figure~\ref{fig:data}). Its 137,827 synthesized and 51,576 adapted examples span 13.52M micro-turns, with 409K \innername entries, 377K \responsename outputs, and 8.6K joint activations preserved from coincident boundaries in the adaptation route. These cases supervise simultaneous branch activation under the same model interface used throughout training and inference.

\subsection{Two-Stage Training}
We first use supervised fine-tuning (SFT) on \ours-189K to teach the model when and how to reason and respond. We then use trajectory-level reinforcement learning (RL) to improve response timing across streaming trajectories.
\subsubsection{Supervised Fine-Tuning}

At each micro-turn, cross-entropy over the branch-specific spaces $\mathcal{A}_I$ and $\mathcal{A}_R$ defines $\mathcal{L}_{\mathrm{action}}$ over the decisions in Equation~\ref{eq:branch_actions} \citep{chen2024videollmonline}. Subsequent tokens from active branches define $\mathcal{L}_{\mathrm{content}}$. To address label imbalance, we upweight $\langle\mathrm{inner}\rangle$ and $\langle\mathrm{speak}\rangle$ separately and downweight all but the first $\langle\mathrm{silent}\rangle$ action in each silent span \citep{yao2026joyai}.

The action loss scores each micro-turn independently, although correct timing requires the response position to dominate nearby silent positions. Action Temporal Ranking (ATR) therefore imposes a local margin at each response boundary \citep{burges2005learning}. Let $z^{R}_{t,a}$ be the \responsename logit for action $a$ at time $t$, and let $r_t=z^{R}_{t,\langle\mathrm{speak}\rangle}-z^{R}_{t,\langle\mathrm{silent}\rangle}$ be its response preference. For a ground-truth response time $t$, let $\mathcal{N}_K(t)$ contain up to $K$ adjacent micro-turns labeled $\langle\mathrm{silent}\rangle$. We select the hardest local negative $u_t=\arg\max_{u\in\mathcal{N}_K(t)}r_u$ and optimize
\begin{equation}
\mathcal{L}_{\mathrm{ATR}}
=
\frac{1}{|\mathcal{P}|}
\sum_{t\in\mathcal{P}}
\operatorname{softplus}\!\left(\delta+r_{u_t}-r_t\right),
\label{eq:atr_loss}
\end{equation}
where $\mathcal{P}$ contains response times with silent neighbors and $\delta$ is the margin. ATR encourages the annotated response preference to exceed that of nearby silent micro-turns. The full objective is
\begin{equation}
\mathcal{L}_{\mathrm{SFT}}
=
\mathcal{L}_{\mathrm{action}}
+\mathcal{L}_{\mathrm{content}}
+\lambda_{\mathrm{ATR}}\mathcal{L}_{\mathrm{ATR}}.
\label{eq:sft_objective}
\end{equation}
Only the two LoRA adapters and action-token parameters are updated. Appendix~\ref{app:implementation} specifies the adapter configuration, action weights, and ATR hyperparameters.

\subsubsection{Trajectory-Level Reinforcement Learning}
Adaptive response timing requires decisions under the model's evolving history. SFT and ATR supervise actions under annotated histories, while inference conditions on the model's own outputs. Early timing errors can therefore affect later decisions. Trajectory-level RL trains on complete causal rollouts and optimizes event coverage while penalizing unmatched responses. SFT provides content supervision, and we evaluate post-RL content quality separately in Table~\ref{tab:sbpo_fine_grained}.

For trajectory $\tau$, we match predicted response times one-to-one with ordered ground-truth windows. Matched predictions are true positives. Unmatched predictions and uncovered windows are false positives and false negatives, respectively. We use their hard-window F1 as the trajectory reward:
\begin{equation}
R(\tau)
=
F_{1}^{\mathrm{hard}}
=
\frac{2\mathrm{TP}}
{2\mathrm{TP}+\mathrm{FP}+\mathrm{FN}}.
\label{eq:trajectory_reward}
\end{equation}

We uniformly sample four groups: single-response trajectories with short or long post-response silence, and multi-response trajectories with sparse or dense events. This balances response frequency and silence duration. For each query, we sample $G$ trajectories $\{\tau_i\}_{i=1}^{G}$ and standardize their rewards within the group to obtain relative advantages for policy optimization:
\begin{equation}
\widehat{A}_i
=
\frac{R(\tau_i)-\overline{R}}
{\operatorname{std}(R(\tau_1),\ldots,R(\tau_G))+\epsilon_A}.
\label{eq:group_advantage}
\end{equation}
Here $\overline{R}$ is the group mean and $\epsilon_A>0$ stabilizes normalization.
For token $y_{i,j}$ generated from causal branch history $h_{i,j}$, define the policy ratio
$\rho_{i,j}(\theta)=\pi_{\theta}(y_{i,j}\mid h_{i,j})/\pi_{\mathrm{old}}(y_{i,j}\mid h_{i,j})$.
We optimize the Group Relative Policy Optimization (GRPO) objective \citep{shao2024deepseekmath}
\begin{equation}
\mathcal{L}_{\mathrm{GRPO}}
=-\frac{1}{G}\sum_{i=1}^{G}\frac{1}{L_i}\sum_{j=1}^{L_i}
\min\!\left(
\rho_{i,j}(\theta)\widehat{A}_i,
\operatorname{clip}\!\left(\rho_{i,j}(\theta),1-\epsilon,1+\epsilon\right)\widehat{A}_i
\right),
\label{eq:grpo_objective}
\end{equation}
where $L_i$ counts the tokens generated by both branches and $\epsilon$ is the clipping radius. The terminal timing advantage weights every generated token in the paired rollout.

\begin{table*}[t]
    \caption{\textbf{Results on StreamingBench and ProactiveVideoQA.} StreamingBench's Proactive Output (PO) uses Acc@2s with a two-second tolerance. ProactiveVideoQA subsets use PAUC ($\omega=0.5$) to evaluate timing and content. Values are percentages.}
    \label{tab:main_proactive}
    \centering
    \small
    \setlength{\tabcolsep}{4pt}
    \renewcommand{\arraystretch}{1.04}
    \begin{tabular*}{\textwidth}{@{\extracolsep{\fill}}lcccc@{}}
        \toprule
        \multirow{2}{*}{\textbf{Method}} & \textbf{StreamingBench} & \multicolumn{3}{c}{\textbf{ProactiveVideoQA}} \\
        \cmidrule(lr){2-2}\cmidrule(l){3-5}
        & \textbf{PO} & \textbf{WEB} & \textbf{EGO} & \textbf{VAD} \\
        \midrule
        \rowcolor{tableblue}\multicolumn{5}{c}{\textbf{Streaming models}} \\
        VideoLLM-online~\citep{chen2024videollmonline} & 4.00 & 25.90 & 25.00 & 25.00 \\
        MMDuet~\citep{wang2025videollm} & 29.44 & 38.90 & \second{46.00} & 27.40 \\
        MMDuet2~\citep{wang2026mmduet2} & 34.60 & \best{53.30} & 33.60 & 28.90 \\
        Em-Garde~\citep{zheng2026garde} & 38.00 & 44.30 & \best{52.30} & 27.40 \\
        VideoChat3~\citep{li2026videochat3} & -- & 28.40 & 28.10 & 25.10 \\
        \midrule
        \rowcolor{tablepurple}\multicolumn{5}{c}{\textbf{Reasoning-enabled models}} \\
        VST~\citep{guan2026video} & 19.60 & 21.11 & 24.97 & 25.08 \\
        ThinkStream~\citep{liu2026thinking} & 27.20 & 30.19 & 24.47 & 18.17 \\
        \midrule
        \rowcolor{tablepeach}\multicolumn{5}{c}{\textbf{Ours}} \\
        \oursrl & \best{41.20} & \second{48.28} & 33.74 & \best{32.43} \\
        \ourssft & \second{38.80} & 47.83 & 33.89 & \second{32.14} \\
        \interleaved & 34.00 & 29.89 & 25.00 & 25.91 \\
        \responseonly & 22.80 & 27.98 & 27.98 & 31.17 \\
        \bottomrule
    \end{tabular*}
    \par\smallskip

    \begin{minipage}{0.98\textwidth}
        \scriptsize
        Best and second-best results are \textbf{bold} and \underline{underlined}. Streaming-model scores follow Em-Garde, except MMDuet's PO and VideoChat3's PVQA results from their papers. We evaluate VST and ThinkStream at 2 fps using released code and original protocols.
    \end{minipage}
\end{table*}

\section{Experiments}
\label{sec_exp}
\subsection{Experimental Setup}
\label{sec:experiments}

\paragraph{Training and inference.}
We initialize \ours from Qwen3.5-4B and train \ourssft on \ours-189K \citep{qwen2026qwen35}. RL makes one pass over 800 unique examples balanced across the four behavior groups. For the proactive evaluations, \oursrl, \ourssft, \interleaved, and \responseonly run at 1 fps. Appendix~\ref{app:implementation} provides the training, decoding, and baseline configurations used in these controlled comparisons.

\paragraph{Comparison methods.}
We compare with VideoLLM-online, Flash-VStream, Dispider, StreamForest, MMDuet, MMDuet2, Em-Garde, and VideoChat3 \citep{chen2024videollmonline,zhang2024flash,qian2025dispider,zeng2026streamforest,wang2025videollm,wang2026mmduet2,zheng2026garde,li2026videochat3}. Scores follow Em-Garde, except MMDuet's PO and VideoChat3's PVQA results from their papers. We evaluate VST and ThinkStream at 2 fps with their released implementations and protocols.

\paragraph{Benchmarks and metrics.}
We follow the official frame-indexed protocols \citep{zheng2026garde}. StreamingBench-PO Acc@2s evaluates timing, while Joint@$K$s also requires containment-matched content under the same predictions and timing rule \citep{lin2026streamingbench}. OVO-FAR covers repetition counting (REC), sequential steps (SSR), and clue sufficiency (CRR). It reports precision, recall, and F1 within two seconds \citep{niu2025ovo}. ProactiveVideoQA PAUC ($\omega=0.5$) integrates GPT-4.1 content scores over valid WEB, EGO, and VAD response windows \citep{wang2025proactivevideoqa,openai2025gpt41}. Appendix~\ref{app:full_results} provides full results.

\subsection{Main Results}

\paragraph{StreamingBench Proactive Output.}
\oursrl reaches 41.2\% Acc@2s, exceeding the highest reported baseline by 3.2 points (Table~\ref{tab:main_proactive}). Trajectory-level RL improves on \ourssft's 38.8\% by 2.4 points. Table~\ref{tab:sbpo_fine_grained} additionally evaluates joint timing and content accuracy.

\paragraph{OVO-Bench Forward Active Responding.}
\oursrl achieves 34.19 mean F1, exceeding the highest reported baseline by 3.20 points, with \ourssft second at 32.37 (Table~\ref{tab:ovo}). It leads external baselines on all three tasks, including 50.00 F1 on REC, 6.84 points above Em-Garde. RL improves SSR and REC but reduces CRR F1, yielding a higher overall mean.

\begin{table*}[t]
    \caption{\textbf{OVO-Bench Forward Active Responding.} Precision, recall, and F1 use a 2 s tolerance. Avg. F1 is the task mean. Values are percentages. Best/second-best values are \textbf{bold}/\underline{underlined}.}
    \label{tab:ovo}
    \centering
    \small
    \setlength{\tabcolsep}{4.3pt}
    \renewcommand{\arraystretch}{1.02}
    \begin{tabular}{@{}lcccc@{}}
        \toprule
        \textbf{Method} & \textbf{CRR (P/R/F1)} & \textbf{SSR (P/R/F1)} & \textbf{REC (P/R/F1)} & \textbf{Avg. F1} \\
        \midrule
        \rowcolor{tableblue}\multicolumn{5}{c}{\textbf{Streaming models}} \\
        VideoLLM-online~\citep{chen2024videollmonline} & 2.08/2.08/2.08 & 6.88/11.07/8.49 & 16.46/7.43/10.24 & 6.93 \\
        Flash-VStream~\citep{zhang2024flash} & 5.21/6.25/5.68 & 9.52/2.46/3.91 & 7.49/3.46/4.73 & 4.77 \\
        StreamForest~\citep{zeng2026streamforest} & 2.08/2.08/2.08 & 13.69/8.02/10.11 & 28.90/30.49/29.67 & 13.95 \\
        MMDuet2~\citep{wang2026mmduet2} & 8.47/47.92/14.40 & -- & \best{42.72}/19.33/26.62 & 20.51 \\
        Em-Garde~\citep{zheng2026garde} & 18.22/47.92/26.40 & \second{14.87}/\second{54.88}/\second{23.40} & \second{39.66}/47.33/43.16 & 30.99 \\
        \midrule
        \rowcolor{tablepurple}\multicolumn{5}{c}{\textbf{Reasoning-enabled models}} \\
        VST~\citep{guan2026video} & 4.65/18.75/7.46 & 3.80/2.86/3.26 & 9.63/\best{95.00}/17.48 & 9.40 \\
        ThinkStream~\citep{liu2026thinking} & 3.09/18.75/5.30 & 3.65/49.60/6.80 & 31.48/23.01/26.58 & 12.89 \\
        \midrule
        \rowcolor{tablepeach}\multicolumn{5}{c}{\textbf{Ours}} \\
        \oursrl & 16.99/\best{64.58}/26.91 & \best{17.50}/48.17/\best{25.67} & 35.43/84.95/\best{50.00} & \best{34.19} \\
        \ourssft & 18.09/\second{62.50}/28.06 & 13.14/\best{57.06}/21.36 & 33.12/\second{85.14}/\second{47.69} & \second{32.37} \\
        \interleaved & \best{23.18}/41.97/\second{29.86} & 2.38/1.59/1.90 & 0.52/2.08/0.83 & 10.87 \\
        \responseonly & \second{20.93}/61.75/\best{31.26} & 4.74/47.22/8.61 & 2.31/18.75/4.11 & 14.66 \\
        \bottomrule
    \end{tabular}
\end{table*}

\paragraph{ProactiveVideoQA.}
\oursrl leads anomaly-triggered VAD at 32.43 PAUC, with \ourssft second at 32.14 (Table~\ref{tab:main_proactive}). It also ranks second on WEB at 48.28 and obtains 33.74 on EGO. EGO's long egocentric procedures require sustained tracking of fine-grained events and repeated updates~\citep{wang2025proactivevideoqa}, which our timing-focused RL objective addresses indirectly.

\subsection{Formulation Comparisons}

We compare three supervised formulations on StreamingBench-PO with the same backbone, source trajectories, and decoding budget. \responseonly retains only \responsename targets. \interleaved uses one LoRA route to select $\langle\mathrm{silent}\rangle$, $\langle\mathrm{inner}\rangle$, or $\langle\mathrm{speak}\rangle$ at each micro-turn. It retains earlier \innername updates but only the \responsename target at joint activations. \ourssft uses independent branch actions and specialized routes. Table~\ref{tab:sbpo_fine_grained} compares these formulations. Section~\ref{sec_ana} examines update timing, role specialization, and synchronous decoding.

\begin{table}[H]
    \caption{\textbf{Formulation comparisons on StreamingBench-PO.} Content metrics use containment or exact matching. @Response evaluates samples with a response. Values are percentages.}
    \label{tab:sbpo_fine_grained}
    \centering
    \footnotesize
    \setlength{\tabcolsep}{6pt}
    \renewcommand{\arraystretch}{1.02}
    \begin{tabular}{@{}lcccccc@{}}
        \toprule
        \textbf{Method}
        & \shortstack{\textbf{Content}\\\textbf{Acc.}}
        & \shortstack{\textbf{Exact Content}\\\textbf{Acc.}}
        & \shortstack{\textbf{Content Acc.}\\\textbf{@Response}}
        & \shortstack{\textbf{Exact Content Acc.}\\\textbf{@Response}}
        & \shortstack{\textbf{Joint}\\\textbf{@1s}}
        & \shortstack{\textbf{Joint}\\\textbf{@2s}} \\
        \midrule
        \oursrl & \best{44.0} & \best{25.6} & \second{79.14} & \best{46.04} & \best{23.2} & \best{33.6} \\
        \ourssft & \second{40.0} & \second{22.8} & \best{79.37} & \second{45.24} & \second{20.8} & \second{30.8} \\
        \interleaved & 39.6 & 6.0 & 59.28 & 8.98 & 10.0 & 17.2 \\
        \responseonly & 24.0 & 10.0 & 50.85 & 21.19 & 6.4 & 10.4 \\
        \bottomrule
    \end{tabular}
\end{table}

\FloatBarrier

\paragraph{Comparison with response-only generation.}
Removing internal-update targets lowers both joint accuracy and conditional content accuracy. Relative to \responseonly, \ourssft raises Joint@2s from 10.4\% to 30.8\% and Content Acc.@Response from 50.85\% to 79.37\%. These gains support internal updates for retaining task-relevant information that later responses can use.

\paragraph{Interleaved streaming reasoning.}
\interleaved preserves internal updates but couples the two roles through one activation sequence. It reaches 59.28\% Content Acc.@Response but only 8.98\% Exact Content Acc.@Response. \ourssft raises exact accuracy to 45.24\%, supporting independent branch decisions and specialized routes for generating complete responses.

\paragraph{Effect of trajectory-level reinforcement learning.}
RL refines the SFT policy using timing rewards. Joint@2s rises from 30.8\% to 33.6\%. Content Acc.@Response remains stable at 79.14\%, indicating preserved content quality under timing-focused optimization.

\begin{figure}[H]
    \centering
    \includegraphics[width=\textwidth]{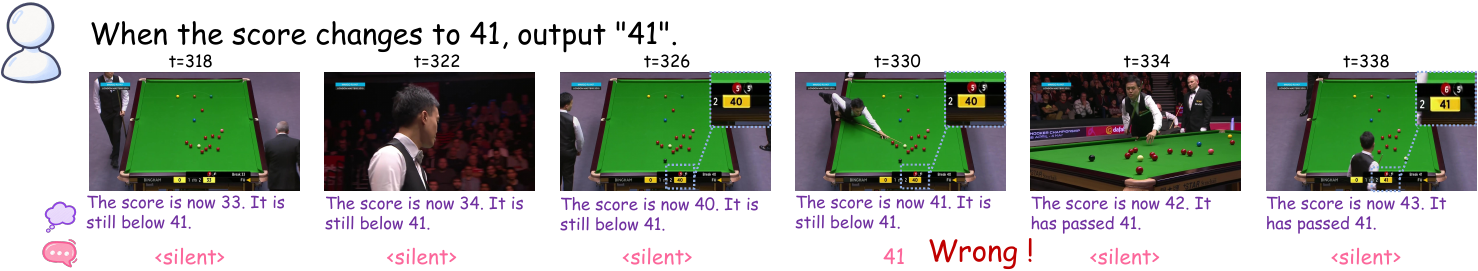}
    \caption{\textbf{A streaming rollout with scheduled \innername activation.}}
    \label{fig:scheduled_case}
\end{figure}

\paragraph{Scheduled reasoning.}
Scheduled uses fixed \innername triggers while \responsename decides autonomously. Figure~\ref{fig:scheduled_case} shows repeated text patterns and redundant \innername updates after the requested response. Adaptive activation can skip redundant updates and focus reasoning on task-relevant changes.

\subsection{Dual-Stream Mechanisms and Capability Retention}
\label{sec_ana}

\begin{figure}[htbp]
    \centering
    \begin{subfigure}[t]{0.425\textwidth}
        \centering
        \includegraphics[width=0.74\linewidth]{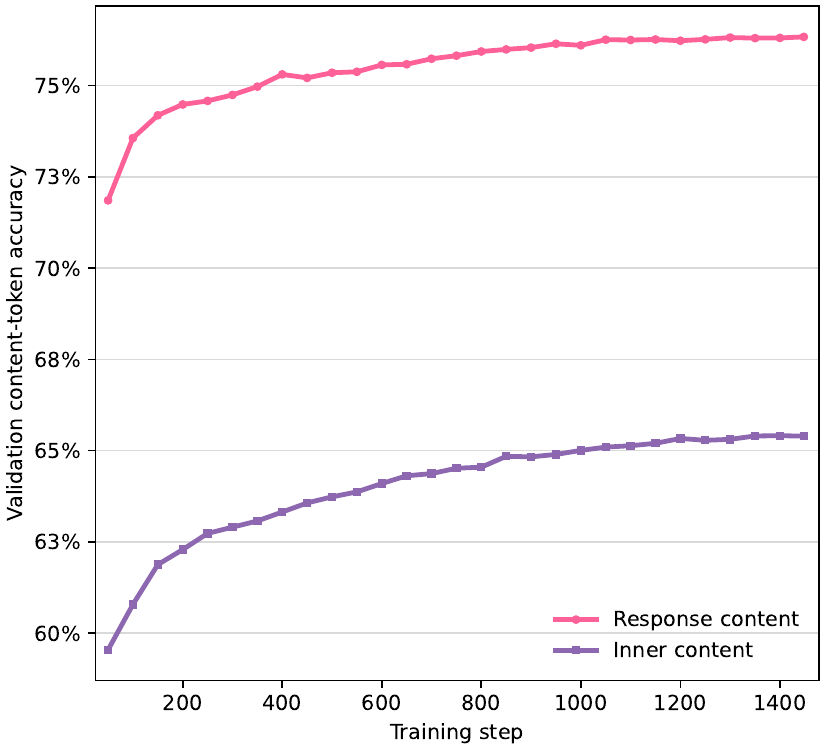}
    \end{subfigure}
    \hfill
    \begin{subfigure}[t]{0.475\textwidth}
        \centering
        \includegraphics[width=0.74\linewidth]{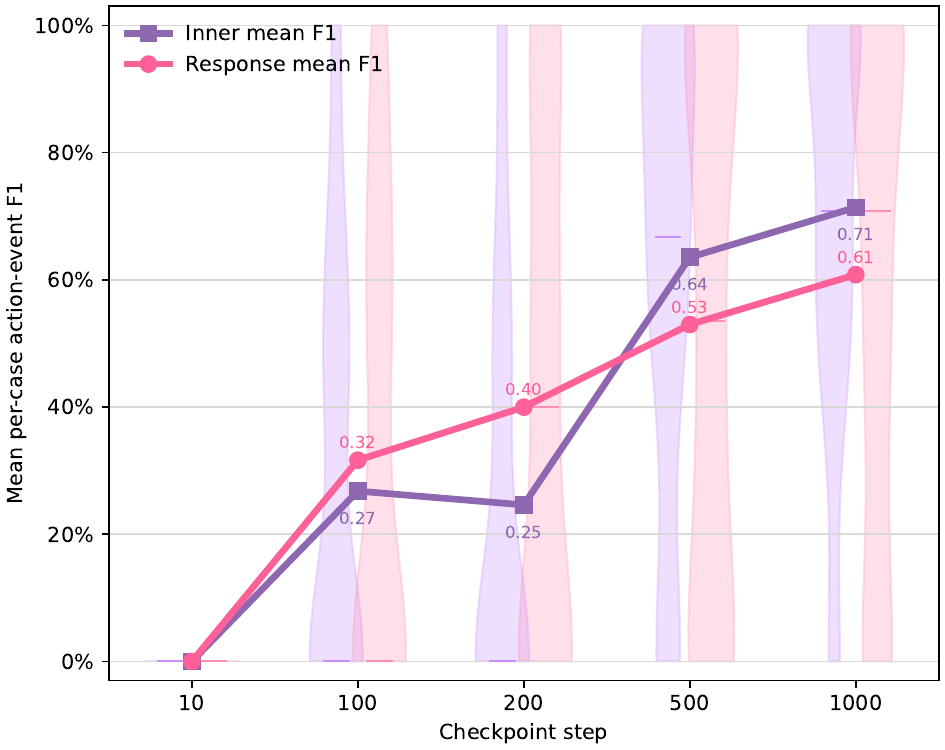}
    \end{subfigure}
    \caption{\textbf{SFT learning dynamics.} Left: validation content-token accuracy over training. Right: mean action-event F1 on 50 evaluated rollouts at five checkpoints.}
    \label{fig:sft_dynamics}
\end{figure}

\paragraph{Temporal alignment of \innername{} updates.}
From step 500 to 1000, \innername activations on 50 rollouts fall from 113 to 93 while exact matches to annotated updates rise from 53 to 56 (Figure~\ref{fig:sft_dynamics}). At step 1000, random update times with the same activation count yield 14.3 exact matches on average, versus 56 for \innername. This supports timing selectivity beyond activation frequency (Appendix~\ref{app:inner_timing}).

\paragraph{Role specialization.}
We swap adapters on 1,870 held-out SFT examples. Reference-token negative log-likelihood rises by 0.706 for \innername and 1.578 for \responsename (Table~\ref{tab:route_swapping}). Assigned adapters better model their own targets, supporting distinct roles for internal updates and responses.

\paragraph{Validation of synchronous decoding.}
Concurrent and \innername-first decoding use the same model, branch conditionals, and attention masks. Outputs agree on all 1,870 held-out examples, preserving generation behavior on these examples. Table~\ref{tab:sync_latency} additionally reports latency for the two execution orders. Appendix~\ref{app:case_studies} illustrates selective updates and separate or joint branch activation.

\begin{table*}[t]
    \centering
    \captionsetup{skip=11pt}
    \begin{minipage}[t]{0.535\textwidth}
        \begin{minipage}[t]{\linewidth}
    \caption{\textbf{General video understanding.}}
    \label{tab:general_video_understanding}
    \centering
    \small
    \setlength{\tabcolsep}{2pt}
    \renewcommand{\arraystretch}{1.5}
    \begin{tabular*}{\linewidth}{@{\extracolsep{\fill}}lccc@{}}
        \toprule
        \textbf{Method} & \textbf{MVBench} & \textbf{EgoSchema} & \textbf{Video-MME} \\
        \midrule
        Qwen3.5-4B & \best{65.42} & \best{61.60} & \best{67.41} \\
        \midrule
        \oursrl & 63.48 & 59.60 & 66.30 \\
        \ourssft & \second{63.55} & 59.60 & 66.41 \\
        \interleaved & 60.82 & \second{60.40} & 64.78 \\
        \responseonly & 61.40 & 59.60 & \second{66.74} \\
        \bottomrule
    \end{tabular*}
\end{minipage}%

    \end{minipage}%
    \hfill%
    \begin{minipage}[t]{0.445\textwidth}
        \begin{minipage}[t]{\linewidth}
    \caption{\textbf{Route specificity.}}
    \label{tab:route_swapping}
    \centering
    \small
    \setlength{\tabcolsep}{2pt}
    \renewcommand{\arraystretch}{1.0}
    \begin{tabular*}{\linewidth}{@{\extracolsep{\fill}}lccc@{}}
        \toprule
        \textbf{Branch} & \textbf{Assigned} & \textbf{Swapped} & \textbf{$\Delta$NLL}$\uparrow$ \\
        \midrule
        \innername & 1.121 & 1.827 & \best{0.706} \\
        \responsename & 0.672 & 2.250 & \best{1.578} \\
        \bottomrule
    \end{tabular*}
\end{minipage}%

        \par\medskip
        \begin{minipage}[t]{\linewidth}
    \caption{\textbf{Decoding latency.}}
    \label{tab:sync_latency}
    \centering
    \small
    \setlength{\tabcolsep}{2pt}
    \renewcommand{\arraystretch}{1.0}
    \begin{tabular*}{\linewidth}{@{\extracolsep{\fill}}lcc@{}}
        \toprule
        \textbf{Metric} & \textbf{Time saved} & \textbf{Speedup} \\
        \midrule
        Both branches finish & 114.74 ms & $1.3706\times$ \\
        First response token & 199.52 ms & $2.2152\times$ \\
        \bottomrule
    \end{tabular*}
\end{minipage}%

    \end{minipage}
\end{table*}

\FloatBarrier

\paragraph{General video understanding.}
Across three offline benchmarks \citep{li2024mvbench,mangalam2023egoschema,fu2025videomme}, both \ours variants remain within 2.0 points of Qwen3.5-4B and retain more MVBench accuracy than \interleaved and \responseonly after proactive adaptation (Table~\ref{tab:general_video_understanding}). The learned streaming policies thus largely preserve general video understanding.

\FloatBarrier

\section{Conclusion}
\label{sec_con}
We introduced \ours{} to coordinate ongoing reasoning with timely responses in streaming video.
Its specialized \innername{} and \responsename{} streams share causal history on one backbone, activate independently, and decode concurrently.
We constructed 189K dual-stream examples and combined supervised fine-tuning with reinforcement learning.
Across three benchmarks, \ours{} improves response timing and content while largely preserving offline video understanding.
These results support explicit reasoning and role separation for proactive video assistants.


\pagebreak

\subsection*{AI use statement}
Gemini 3.1 Pro Preview~\citep{google2026gemini31},
Qwen-VL-Max-Latest~\citep{alibaba2024qwenvlmax}, and
GPT-5.4~\citep{openai2026gpt54} were used for the
synthetic-data pipeline described in Section~\ref{sec_data} and
Appendix~\ref{app:prompts}. Their roles included proactive-query generation,
dual-stream trajectory synthesis and revision, response reconstruction, and
semantic judging. AI tools were also used for language editing. The
authors designed the pipeline and conducted the experiments. They verified the
reported claims, results, and citations, and retain full responsibility for all
scientific decisions and the final manuscript.

\subsection*{Ethics statement}
This work collects no new human-subject data. All experiments use established
public research datasets under their released terms. We identify no additional
ethical concerns beyond those associated with the source datasets and proactive
video systems. Deployment in privacy-sensitive settings requires appropriate
consent, access control, and bias assessment.

\subsection*{Reproducibility statement}
Section~\ref{sec_method} specifies the model, objectives, and
data pipeline. Section~\ref{sec_exp} defines the evaluation protocol, and
Appendix~\ref{app:implementation} provides implementation details. Abridged
construction prompts appear in Appendix~\ref{app:prompts}. We will release the
complete prompts, data, training and evaluation code, configurations, and model
checkpoints required to reproduce the reported experiments.



\bibliography{ref}
\bibliographystyle{iclr2027_conference}

\newpage
\appendix
\raggedbottom
\section{Implementation Details}
\label{app:implementation}

\paragraph{Supervised fine-tuning.}
The vision encoder and shared Qwen3.5-4B backbone remain frozen. Each branch uses a LoRA adapter of rank 64, scaling factor 128, and dropout 0.05. We additionally optimize the vocabulary parameters for the new action tokens. Training uses AdamW with learning rate $10^{-4}$ and global batch size 128 \citep{loshchilov2019decoupled}. The action-token weights are $w_{\mathrm{inner}}=2.5$ and $w_{\mathrm{speak}}=3.0$. Within each consecutive silent span, the first silent action has weight 1.0 and later silent actions have weight 0.2. ATR uses $\lambda_{\mathrm{ATR}}=0.05$, margin $\delta=0.5$, and neighborhood size $K=4$. We train for one epoch using bfloat16 and DeepSpeed ZeRO-2 for distributed optimization \citep{rajbhandari2020zero}.

\paragraph{SFT learning dynamics.}
SFT improves both content generation and action prediction (Figure~\ref{fig:sft_dynamics}). Validation content-token accuracy rises by 5.86 points for \innername and 4.48 for \responsename. Their mean per-case action-event F1 reaches 0.71 and 0.61 on the 50 evaluated rollouts.

\paragraph{Instantiation on the Qwen3.5 backbone.}
Qwen3.5 inherits Qwen3-Next's hybrid decoder, whose linear-attention layers use Gated DeltaNet \citep{qwen2026qwen35,yang2025gated}. At position $j$, the recurrent state follows
\begin{equation}
\widetilde{S}_{j}
=
\alpha_j\widetilde{S}_{j-1}
\left(I-\beta_j k_jk_j^{\top}\right)
+\beta_jv_jk_j^{\top},
\qquad
h_j=\widetilde{S}_{j}q_j,
\label{eq:gated_delta_update}
\end{equation}
where $q_j$, $k_j$, and $v_j$ are the query, key, and value vectors. The data-dependent coefficients $\alpha_j=\exp(g_j)\in(0,1)$ and $\beta_j=\sigma(b_j)\in(0,1)$ control decay and write strength.

At each micro-turn, the teacher-forced \innername and \responsename targets occupy one physical sequence in that order. A block-causal mask lets each branch attend to the shared prefix and its own earlier tokens, while blocking attention between the two current branches. Both completed outputs enter the state exposed to the next micro-turn. Full-attention layers construct the two KV increments under this mask. Linear-attention layers fork the shared recurrent state, process both targets independently, and apply Equation~\ref{eq:linear_attention_merge} before the next micro-turn. Training and inference use the same token order, mask, position IDs, and state transition. Inference replaces teacher-forced targets with tokens generated autoregressively by each branch.

\paragraph{State merging and positions.}
Both branches represent simultaneous decisions from one shared prefix. We set $\gamma=0.5$ as a symmetric heuristic that assigns neither role prior weight. The resulting mean $(S_t^I+S_t^R)/2$ is invariant to branch order and generally differs from either serial composition. Both LoRA routes are trained under this rule, which is applied unchanged at inference. For three-axis MRoPE, both branches start at the first position after the prefix to encode a common temporal origin, following position-sharing conventions for parallel decoding and aligned multimodal streams \citep{chen2025aspd,xueyun2026roma}. The next micro-turn advances by $\max(|o_t^I|,|o_t^R|)$, placing its first position after the longer branch span.

\paragraph{Merge-weight parameterization.}
We compare two merge-weight settings under the same SFT configuration (Table~\ref{tab:merge_weights}). Fixed uses $\gamma=0.5$ and corresponds to the reported \ourssft model. Learned per branch optimizes separate coefficients for \innername and \responsename.

\FloatBarrier
\begin{table*}[t]
    \caption{\textbf{State-merge weights.} (a) StreamingBench-PO Acc@2s. (b) OVO-FAR precision, recall, and F1 within two seconds. Fixed uses $\gamma=0.5$. Learned per branch optimizes separate coefficients for \innername and \responsename. Values are percentages.}
    \label{tab:merge_weights}
    \centering
    \captionsetup[subtable]{font=small,labelfont=bf,skip=4pt}
    \begin{subtable}[t]{0.30\textwidth}
        \caption{StreamingBench-PO.}
        \label{tab:merge_weights_sbpo}
        \centering
        \footnotesize
        \setlength{\tabcolsep}{3.0pt}
        \renewcommand{\arraystretch}{1.08}
        \begin{tabular}{@{}lc@{}}
            \toprule
            \textbf{Merge weights} & \textbf{Acc@2s} \\
            \midrule
            Fixed & 38.8 \\
            Learned per branch & 38.0 \\
            \bottomrule
        \end{tabular}
    \end{subtable}
    \hfill
    \begin{subtable}[t]{0.68\textwidth}
        \caption{OVO-FAR.}
        \label{tab:merge_weights_ovo}
        \centering
        \fontsize{7}{7.8}\selectfont
        \setlength{\tabcolsep}{2.5pt}
        \renewcommand{\arraystretch}{1.08}
        \begin{tabular}{@{}lcccc@{}}
            \toprule
            \textbf{Merge weights}
            & \textbf{CRR (P/R/F1)}
            & \textbf{SSR (P/R/F1)}
            & \textbf{REC (P/R/F1)}
            & \textbf{Avg. F1} \\
            \midrule
            Fixed & 18.09/62.50/28.06 & 13.14/57.06/21.36 & 33.12/85.14/47.69 & 32.37 \\
            Learned per branch & 13.91/60.42/22.61 & 11.51/49.84/18.70 & 36.32/84.63/50.83 & 30.71 \\
            \bottomrule
        \end{tabular}
    \end{subtable}
\end{table*}

\paragraph{Effect of merge weights.}
Learning separate branch weights reduces StreamingBench-PO Acc@2s from 38.8\% to 38.0\% and OVO-FAR mean F1 from 32.37\% to 30.71\% (Table~\ref{tab:merge_weights}). The OVO-FAR results show a task-dependent trade-off: REC F1 increases by 3.14 points, while CRR and SSR F1 decrease by 5.45 and 2.66 points. The REC gain comes from higher precision, whereas both precision and recall decline on CRR and SSR. Fixed symmetric merging thus performs better overall than independent branch weighting under the evaluated SFT configuration.

\paragraph{Reinforcement learning.}
RL starts from the SFT checkpoint and updates only the two branch-specific LoRA adapters. The reward contains only the timing term. In pilot runs, adding a content term increased response frequency and verbosity without improving conditional content accuracy. We use AdamW with FP32 master weights, $\beta_1=0.9$, $\beta_2=0.999$, $\epsilon=10^{-8}$, learning rate $10^{-6}$, zero weight decay, and gradient clipping at 1.0. GRPO uses a clipping radius of 0.2 and a zero KL coefficient. RL makes one pass over 800 unique examples, with 200 from each of four behavior groups: single-response trajectories with short or long post-response silence, and multi-response trajectories with sparse or dense response events. Each of the 50 updates samples 16 queries without replacement, four per group, and draws $G=4$ rollouts per query, yielding 64 trajectories. The reported model is the checkpoint after the final update.

\paragraph{Streaming inference.}
We implement a Qwen3.5 streaming runtime with Hugging Face Transformers and PyTorch FlexAttention \citep{qwen2026qwen35,dong2025flexattention}. Videos are sampled at 1 fps, and each frame is encoded once into at most 128 visual tokens. At each micro-turn, the runtime forks the shared state, applies both LoRA routes, and batches their autoregressive decoding. Both branches use greedy decoding with a limit of 256 generated tokens per branch and micro-turn. Their states are merged before the next frame, while only active \responsename{} content is streamed to the user during decoding.

For adaptive inference, action tokens determine activation directly. No response threshold is tuned on the evaluation sets. Benchmark decisions are indexed by sampled frames, which the runner processes in order without dropping or coalescing. The runtime reuses the KV cache and evicts frame tokens after their state update. Full-attention KV memory still grows with retained action and text tokens. Evicting frame tokens removes the dominant per-frame contribution and limits persistent growth to compact branch outputs retained across micro-turns.

\paragraph{Scheduling semantics.}
The scheduling comparison executes one fixed \ours model in two ways. Concurrent execution batches both branches from the shared state. The \innername-first schedule completes \innername before decoding \responsename from the same state, without introducing cross-branch conditioning. The two schedules therefore preserve branch outputs and the merged transition while changing only execution order. Time to the first \responsename token includes preceding \innername computation under the sequential schedule. Task scores remain frame-indexed, while Appendix~\ref{app:runtime_profile} reports wall-clock service times for one stream under the stated runtime configuration.

\paragraph{Adapted baselines.}
For the proactive comparisons in Tables~\ref{tab:main_proactive} and~\ref{tab:ovo}, we evaluate VST and ThinkStream from their released implementations and checkpoints at 2 fps, following each method's inference protocol \citep{guan2026video,liu2026thinking}. Their emitted answers are treated as response events, while steps without an answer remain silent. The broader task results in Tables~\ref{tab:streamingbench_full} and~\ref{tab:ovobench_full} use the listed frame rates. Other streaming-model values follow the Em-Garde compilation \citep{zheng2026garde}, except MMDuet's StreamingBench-PO and VideoChat3's ProactiveVideoQA scores from their original reports \citep{wang2025videollm,li2026videochat3}. Their frame rates and model sizes follow the reported sources. Controlled formulation comparisons are confined to our four variants under the common 1-fps protocol and matched decoding budget.

\paragraph{Data provenance and generative tools.}
All examples in \ours-189K are constructed from the official training splits of the source datasets, while StreamingBench, OVO-Bench, and ProactiveVideoQA are reserved for evaluation. For the UCF-Crime source also used by ProactiveVideoQA, the training and held-out test videos have disjoint IDs \citep{wang2025proactivevideoqa}. To preserve the streaming setting, each \innername or \responsename target at timestamp $t$ is generated from the query, preceding trajectory entries, and a 1-fps sample of $\mathcal{V}_{\leq t}$. The complete video is used only for trajectory-level auditing of event coverage and trigger timing, and candidates selected for refinement are regenerated under the same prefix constraint. Gemini 3.1 Pro Preview is used for proactive query generation and judging, Qwen-VL-Max-Latest for dual-stream trajectory synthesis, and GPT-5.4 for targeted refinement. Appendix~\ref{app:prompts} details the interfaces and acceptance criteria for each stage.

\section{Long-Stream Runtime Profile}
\label{app:runtime_profile}

We profile \ourssft in BF16 over 480 seconds of a StreamingBench video sampled at 1 fps. The source resolution is $1920\times1080$. Each sampled frame is resized to $480\times256$ and encoded as 120 visual tokens. The run uses one video stream, greedy decoding with at most 64 tokens per branch, a $W=1$ step mask, and a 16,384-token preallocated streaming cache. The mask evicts prior visual KV while retaining the system prompt, query, action tokens, and generated text. End-to-end step time includes image preprocessing, visual encoding, shared prefill, batched dual-branch decoding, state merging, and cache eviction. It excludes disk I/O and video decoding. We omit the first two steps affected by compilation. The trace measures per-step service time and excludes queueing across concurrent streams, isolating the execution cost of one video stream.

\begin{figure}[htbp]
    \centering
    \begin{subfigure}[t]{0.242\textwidth}
        \centering
        \includegraphics[width=\linewidth]{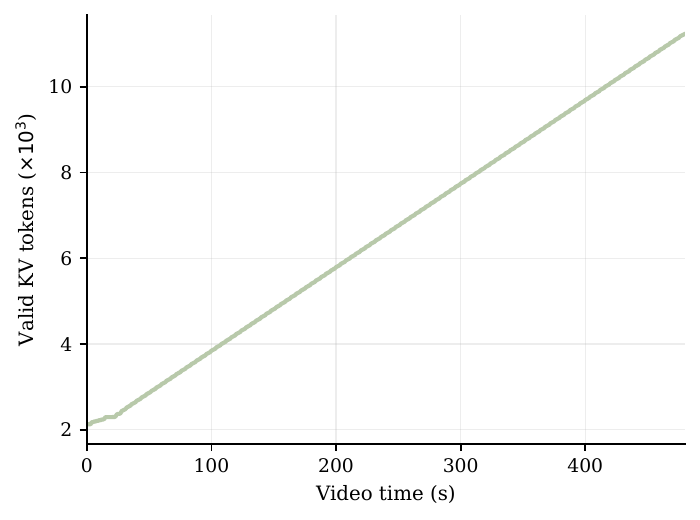}
        \caption{Retained KV length.}
        \label{fig:runtime_kv}
    \end{subfigure}
    \hfill
    \begin{subfigure}[t]{0.242\textwidth}
        \centering
        \includegraphics[width=\linewidth]{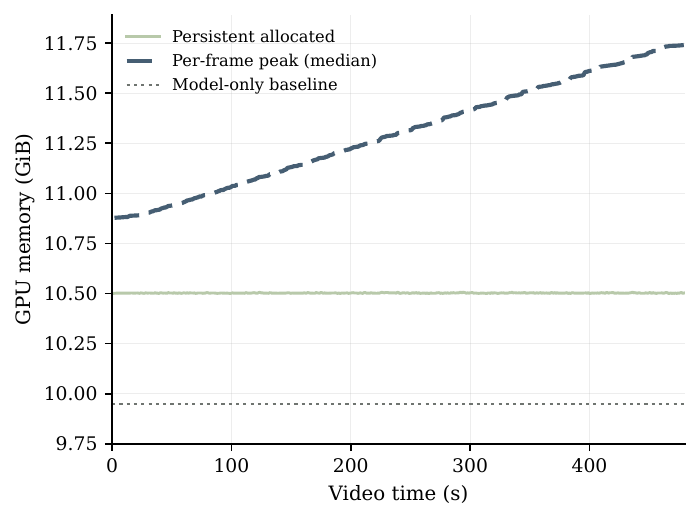}
        \caption{GPU memory.}
        \label{fig:runtime_memory}
    \end{subfigure}
    \hfill
    \begin{subfigure}[t]{0.242\textwidth}
        \centering
        \includegraphics[width=\linewidth]{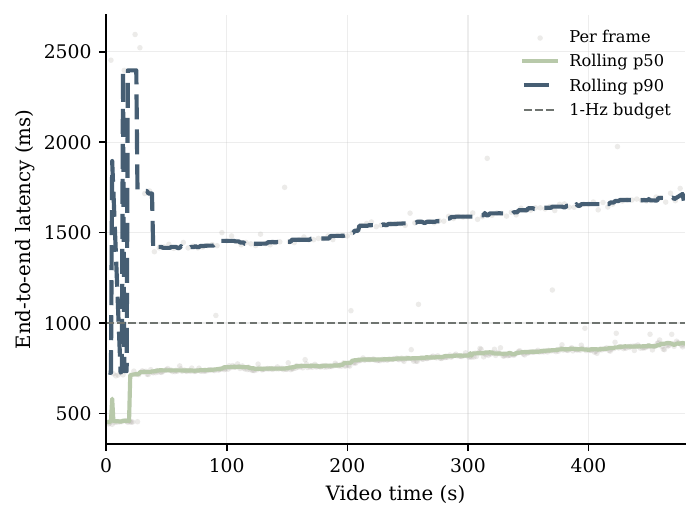}
        \caption{End-to-end latency.}
        \label{fig:runtime_e2e}
    \end{subfigure}
    \hfill
    \begin{subfigure}[t]{0.242\textwidth}
        \centering
        \includegraphics[width=\linewidth]{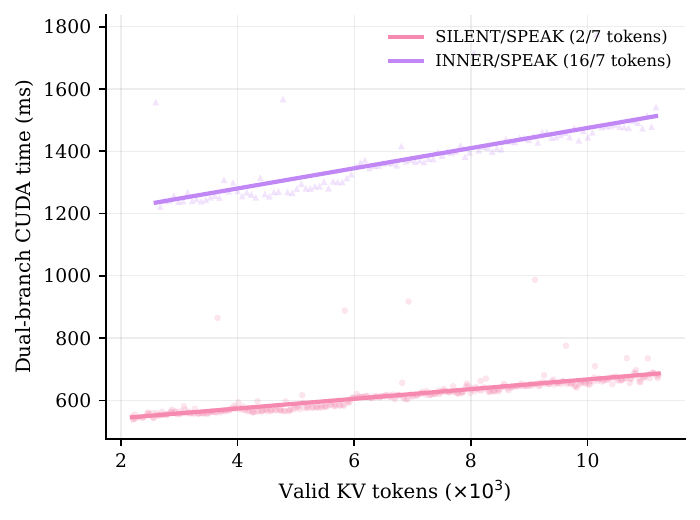}
        \caption{Branch latency.}
        \label{fig:runtime_branch}
    \end{subfigure}
    \caption{\textbf{Long-stream memory and latency of dual-branch inference.} We profile a 480-s stream at 1 fps, excluding the first two compilation-affected steps. Fixed-capacity cache allocation keeps persistent memory stable, while the growing retained prefix increases transient peak memory and decoding cost. Panel (d) fixes branch output lengths to isolate how retained KV context length affects decoding latency for each branch activation pattern.}
    \label{fig:long_stream_runtime}
\end{figure}
\FloatBarrier

Figure~\ref{fig:long_stream_runtime} shows that retained full-attention KV grows from 2,126 to 11,220 tokens. Persistent allocation changes by only 1.1 MiB, while the per-frame peak rises by 0.88 GiB to 11.75 GiB. Across 478 measured steps, mean and median end-to-end latency are 985.3 and 827.2 ms. The p90 is 1,596.7 ms, and 74.9\% of frames meet the 1-s budget. Panel~\ref{fig:runtime_branch} fixes branch output lengths within each action pair. Dual-branch CUDA latency increases by 15.6 and 32.4 ms per 1k retained KV tokens when only \responsename is active and when both branches are active, respectively. In this workload, cache preallocation stabilizes persistent allocation, while retained text increases transient memory and long-stream latency over the full 480-s stream.

\section{Detailed Benchmark Results}
\label{app:full_results}

Tables~\ref{tab:streamingbench_full} and~\ref{tab:ovobench_full} provide task-level results and frame metadata for the broader benchmark suites \citep{lin2026streamingbench,niu2025ovo}. Streaming baselines use the compilation of Em-Garde, while VST and ThinkStream are our 2-fps evaluations \citep{zheng2026garde,guan2026video,liu2026thinking}.

On StreamingBench, RL improves SFT on six of ten tasks. The largest gains occur on EU and CR, at 1.89 and 1.56 points, while the other four task scores remain unchanged. The resulting overall score rises from 70.44 to 70.84 after trajectory-level optimization.

On OVO-Bench, both variants attain the strongest online FPD score of 76.24, while \ourssft leads online models on ASI at 68.24. RL maintains average accuracy on the Real-Time Visual Perception subset (61.17 vs. 61.13 for SFT) and improves STU by 1.12 points. OCR, ACR, FPD, and OJR remain unchanged. The backward-tracing average shifts slightly from 43.95 to 43.62.

\begin{table*}[b]
    \caption{\textbf{StreamingBench Real-Time Visual Understanding subset.} Bold and underlined values mark the best and second-best results among open-source online models.}
    \label{tab:streamingbench_full}
    \centering
    \scriptsize
    \setlength{\tabcolsep}{2.8pt}
    \renewcommand{\arraystretch}{1.04}
    \resizebox{\textwidth}{!}{%
    \begin{tabular}{@{}lccccccccccccc@{}}
        \toprule
        \textbf{Method} & \textbf{Size} & \textbf{Frames} & \textbf{OP} & \textbf{CR} & \textbf{CS} & \textbf{ATP} & \textbf{EU} & \textbf{TR} & \textbf{PR} & \textbf{SU} & \textbf{ACP} & \textbf{CT} & \textbf{Overall} \\
        \midrule
        Human & -- & -- & 89.50 & 92.00 & 93.60 & 91.50 & 95.70 & 92.50 & 88.00 & 88.80 & 89.70 & 91.30 & 91.50 \\
        \midrule
        \rowcolor{tablepurple}\multicolumn{14}{c}{\textbf{Proprietary models}} \\
        Gemini 1.5 Pro    & -- & 1 fps & 79.00 & 80.50 & 83.50 & 79.70 & 80.00 & 84.70 & 77.80 & 64.20 & 72.00 & 48.70 & 75.70 \\
        GPT-4o            & -- & 64    & 77.10 & 80.50 & 83.90 & 76.50 & 70.20 & 83.80 & 66.70 & 62.20 & 69.10 & 49.20 & 73.30 \\
        Claude 3.5 Sonnet & -- & 20    & 73.30 & 80.50 & 84.10 & 82.00 & 75.40 & 79.50 & 61.10 & 61.80 & 69.30 & 43.10 & 72.40 \\
        \midrule
        \rowcolor{tablepurple}\multicolumn{14}{c}{\textbf{Offline models}} \\
        Video-LLaMA2     & 7B  & 32    & 55.90 & 55.50 & 57.40 & 58.20 & 52.80 & 43.60 & 39.80 & 42.70 & 45.60 & 35.20 & 49.50 \\
        VILA-1.5         & 8B  & 14    & 53.70 & 49.20 & 71.00 & 56.90 & 53.40 & 53.90 & 54.60 & 48.80 & 50.10 & 17.60 & 52.30 \\
        Video-CCAM       & 14B & 96    & 56.40 & 57.80 & 65.30 & 62.80 & 64.60 & 51.40 & 42.60 & 48.00 & 49.60 & 31.60 & 54.00 \\
        LongVA           & 7B  & 128   & 70.00 & 63.30 & 61.20 & 70.90 & 62.70 & 59.50 & 61.10 & 53.70 & 54.70 & 34.70 & 60.00 \\
        InternVL2        & 8B  & 16    & 68.10 & 60.90 & 69.40 & 77.10 & 67.70 & 62.90 & 59.30 & 53.30 & 55.00 & 56.50 & 63.70 \\
        Kangaroo         & 7B  & 64    & 71.10 & 84.40 & 70.70 & 73.20 & 67.10 & 61.70 & 56.50 & 55.70 & 62.00 & 38.90 & 64.60 \\
        LLaVA-NeXT-Video & 32B & 64    & 78.20 & 70.30 & 73.80 & 76.80 & 63.40 & 69.80 & 57.40 & 56.10 & 64.30 & 38.90 & 67.00 \\
        MiniCPM-V2.6     & 8B  & 32    & 71.90 & 71.10 & 77.90 & 75.80 & 64.60 & 65.70 & 70.40 & 56.10 & 62.30 & 53.40 & 67.40 \\
        LLaVA-OneVision  & 7B  & 32    & 80.40 & 74.20 & 76.00 & 80.70 & 72.70 & 71.70 & 67.60 & 65.50 & 65.70 & 45.10 & 71.10 \\
        Qwen2.5-VL       & 7B  & 1 fps & 78.30 & 80.50 & 78.90 & 80.50 & 76.70 & 78.50 & 79.60 & 63.40 & 66.20 & 53.20 & 73.70 \\
        \midrule
        \rowcolor{tableblue}\multicolumn{14}{c}{\textbf{Streaming models}} \\
        Flash-VStream   & 7B & --    & 25.90 & 43.60 & 24.90 & 23.90 & 27.30 & 13.10 & 18.50 & 25.20 & 23.90 & 48.70 & 23.20 \\
        VideoLLM-online & 8B & 2 fps & 39.10 & 40.10 & 34.50 & 31.10 & 46.00 & 32.40 & 31.50 & 34.20 & 42.50 & 27.90 & 36.00 \\
        Dispider        & 7B & 1 fps & 74.90 & 75.50 & 74.10 & 73.10 & 74.40 & 59.90 & 76.10 & 62.90 & 62.20 & 45.80 & 67.60 \\
        StreamAgent     & 7B & 1 fps & 79.60 & 78.30 & 79.20 & 75.90 & 74.70 & 76.90 & \second{82.90} & 66.30 & 73.70 & \best{55.40} & 74.30 \\
        StreamForest    & 7B & 1 fps & 83.10 & \best{82.80} & 82.70 & \second{84.30} & \second{77.50} & 78.20 & 76.90 & 69.10 & \best{75.60} & \second{54.40} & \second{77.30} \\
        Em-Garde        & 7B & 2 fps & 80.40 & 74.20 & \best{89.90} & 84.00 & \best{78.90} & \second{82.20} & \best{87.00} & \second{71.10} & 68.80 & 42.50 & 76.70 \\
        \midrule
        \rowcolor{tablepurple}\multicolumn{14}{c}{\textbf{Reasoning-enabled models}} \\
        VST~\citep{guan2026video}           & 7B & 2 fps & \best{85.40} & \second{82.00} & \second{86.40} & \best{89.10} & 74.20 & \best{87.20} & 82.40 & \best{73.10} & 73.90 & 47.30 & \best{79.50} \\
        ThinkStream~\citep{liu2026thinking} & 3B & 2 fps & \second{83.74} & 70.31 & 76.03 & 82.69 & 76.73 & 78.19 & 76.85 & 70.73 & \second{74.43} & 45.21 & 75.00 \\
        \midrule
        \rowcolor{tablepeach}\multicolumn{14}{c}{\textbf{Ours}} \\
        \oursrl  & 4B & 1 fps & 66.94 & 81.25 & 77.29 & 80.77 & 70.44 & 76.32 & 75.00 & 60.16 & 67.90 & 52.13 & 70.84 \\
        \ourssft & 4B & 1 fps & 66.67 & 79.69 & 77.29 & 80.45 & 68.55 & 76.32 & 75.00 & 59.76 & 67.90 & 51.06 & 70.44 \\
        \bottomrule
    \end{tabular}%
    }
\end{table*}

\begin{table*}[t]
    \caption{\textbf{OVO-Bench task accuracy.} Bold and underlined values mark the best and second-best results among open-source online models.}
    \label{tab:ovobench_full}
    \centering
    \scriptsize
    \setlength{\tabcolsep}{3.0pt}
    \renewcommand{\arraystretch}{1.05}
    \resizebox{\textwidth}{!}{%
    \begin{tabular}{@{}lccccccccccccc@{}}
        \toprule
        \multirow{2}{*}{\textbf{Method}} & \multirow{2}{*}{\textbf{Size}} & \multirow{2}{*}{\textbf{Frames}} &
        \multicolumn{7}{c}{\textbf{Real-Time Visual Perception}} &
        \multicolumn{4}{c}{\textbf{Backward Tracing}} \\
        \cmidrule(lr){4-10}\cmidrule(lr){11-14}
        & & & \textbf{OCR} & \textbf{ACR} & \textbf{ATR} & \textbf{STU} & \textbf{FPD} & \textbf{OJR} & \textbf{Avg.} & \textbf{EPM} & \textbf{ASI} & \textbf{HLD} & \textbf{Avg.} \\
        \midrule
        \rowcolor{tablepurple}\multicolumn{14}{c}{\textbf{Proprietary models}} \\
        Gemini 1.5 Pro & -- & -- & 85.90 & 67.00 & 79.30 & 58.40 & 63.40 & 62.00 & 69.30 & 58.60 & 76.40 & 52.60 & 62.50 \\
        GPT-4o         & -- & -- & 69.80 & 64.20 & 71.60 & 51.10 & 70.30 & 59.80 & 64.50 & 57.90 & 75.70 & 48.70 & 60.80 \\
        \midrule
        \rowcolor{tablepurple}\multicolumn{14}{c}{\textbf{Offline models}} \\
        LLaVA-Video     & 7B & 64    & 69.80 & 59.60 & 66.40 & 50.60 & 72.30 & 61.40 & 63.50 & 51.20 & 64.20 & 9.70  & 41.70 \\
        LLaVA-OneVision & 7B & 64    & 67.10 & 58.70 & 69.80 & 49.40 & 71.30 & 60.30 & 62.80 & 52.50 & 58.80 & 23.70 & 45.00 \\
        Qwen2-VL        & 7B & 64    & 69.10 & 53.20 & 63.80 & 50.60 & 66.30 & 60.90 & 60.70 & 44.40 & 66.90 & 34.40 & 48.60 \\
        InternVL-V2     & 8B & 64    & 68.50 & 58.70 & 69.00 & 44.90 & 67.30 & 56.00 & 60.70 & 43.10 & 61.50 & 27.40 & 44.00 \\
        LongVU          & 7B & 1 fps & 55.70 & 49.50 & 59.50 & 48.30 & 68.30 & 63.00 & 57.40 & 43.10 & 66.20 & 9.10  & 39.50 \\
        \midrule
        \rowcolor{tableblue}\multicolumn{14}{c}{\textbf{Streaming models}} \\
        Flash-VStream   & 7B & 1 fps & 25.50 & 32.10 & 29.30 & 33.70 & 29.70 & 28.80 & 29.90 & 36.40 & 33.80 & 5.90  & 25.40 \\
        VideoLLM-online & 8B & 2 fps & 8.10  & 23.90 & 12.10 & 14.00 & 45.50 & 21.20 & 20.80 & 22.20 & 18.80 & 12.20 & 17.70 \\
        Dispider        & 7B & 1 fps & 57.70 & 49.50 & 62.10 & 44.90 & 61.40 & 51.60 & 54.50 & 48.50 & 55.40 & 4.30  & 36.10 \\
        TimeChat-Online & 7B & 1 fps & 69.80 & 48.60 & 64.70 & 44.90 & 68.30 & 55.40 & 58.60 & 53.90 & 62.80 & 9.10  & 42.00 \\
        StreamAgent     & 7B & 1 fps & 71.20 & 53.20 & 63.60 & \second{53.90} & 67.30 & 58.70 & 61.30 & 54.80 & 58.10 & 25.80 & 41.70 \\
        \midrule
        \rowcolor{tablepurple}\multicolumn{14}{c}{\textbf{Reasoning-enabled models}} \\
        VST~\citep{guan2026video}           & 7B & 2 fps & \second{80.50} & \second{55.10} & \best{72.40} & \best{55.10} & \second{76.20} & \second{64.10} & \best{67.20} & \best{56.90} & 64.90 & \best{48.40} & \best{56.70} \\
        ThinkStream~\citep{liu2026thinking} & 3B & 2 fps & \best{85.23} & \best{64.22} & \second{69.82} & 49.43 & 69.31 & \best{64.13} & \second{67.03} & 53.87 & 59.46 & \second{43.55} & \second{52.30} \\
        \midrule
        \rowcolor{tablepeach}\multicolumn{14}{c}{\textbf{Ours}} \\
        \oursrl  & 4B & 1 fps & 69.80 & 53.21 & 67.24 & 50.00 & \best{76.24} & 50.54 & 61.17 & 55.22 & \second{67.57} & 8.06 & 43.62 \\
        \ourssft & 4B & 1 fps & 69.80 & 53.21 & 68.10 & 48.88 & \best{76.24} & 50.54 & 61.13 & \second{55.56} & \best{68.24} & 8.06 & 43.95 \\
        \bottomrule
    \end{tabular}%
    }
\end{table*}

\clearpage
\section{Analysis of Inner Update Timing}
\label{app:inner_timing}

We analyze the 50 evaluated rollouts in Figure~\ref{fig:sft_dynamics}, which contain 79 annotated \innername updates. Predicted activations are matched one-to-one with reference updates within each case. Exact matching requires the same micro-turn. Tolerant matching permits a displacement of up to one or two micro-turns. We pool event counts across cases to compute precision, recall, and F1 in Table~\ref{tab:inner_timing}. Figure~\ref{fig:sft_dynamics} reports the mean of per-case F1 scores.

\begin{table}[htbp]
    \centering
    \caption{\textbf{Temporal alignment of \innername updates on 50 rollouts.} All settings use the same 79 reference updates. Precision, recall, and F1 pool event counts across cases and are percentages. Tolerance is measured in micro-turns.}
    \label{tab:inner_timing}
    \small
    \setlength{\tabcolsep}{5pt}
    \begin{tabular}{@{}rrrrrrr@{}}
        \toprule
        \textbf{Step} & \textbf{Tolerance} & \textbf{Activations} & \textbf{Matches} & \textbf{Precision} & \textbf{Recall} & \textbf{F1} \\
        \midrule
        100  & 0 & 101 & 34 & 33.7 & 43.0 & 37.8 \\
        200  & 0 & 75  & 28 & 37.3 & 35.4 & 36.4 \\
        500  & 0 & 113 & 53 & 46.9 & 67.1 & 55.2 \\
        1000 & 0 & 93  & 56 & 60.2 & 70.9 & 65.1 \\
        \midrule
        1000 & 1 & 93 & 68 & 73.1 & 86.1 & 79.1 \\
        1000 & 2 & 93 & 69 & 74.2 & 87.3 & 80.2 \\
        \bottomrule
    \end{tabular}
\end{table}

\paragraph{Learning to select update times.}
From step 500 to 1000, the number of activations falls from 113 to 93 while exact matches increase from 53 to 56. Precision improves by 13.3 percentage points and recall by 3.8 points. The later checkpoint achieves better alignment with fewer activations.

\paragraph{Temporal tolerance.}
At step 1000, allowing one micro-turn of tolerance increases matches from 56 to 68. A second micro-turn adds one match. Among the 69 matches within two micro-turns, 56 are exact, eight occur one step early, four occur one step late, and one occurs two steps early. Thus, most additional matches recovered by temporal tolerance lie within one neighboring micro-turn.

\paragraph{Random timing with matched activation counts.}
We compare the step-1000 activations with random update times under the same activation count. The random baseline averages 14.3 exact matches and 33.2 matches within one micro-turn. The model obtains 56 and 68 matches under the corresponding criteria. This comparison supports temporal alignment with the annotated updates beyond that obtained from random timing at the same activation count.

\clearpage
\section{Data-Construction Prompts}
\label{app:prompts}

The following boxes are formatting-normalized abridgements of the execution prompts. Bracketed ellipses mark omitted explanations and few-shot examples. Each box is kept intact to prevent a prompt from spanning pages. The released artifacts will include the complete execution prompts.

\begin{promptbox}{A1. Dual-stream model interface (abridged)}
ROLE
You are a streaming-video assistant performing proactive response.

INPUT
The user first provides "Query: ...". Video frames then arrive at 1 fps.

At every chunk, make two independent decisions:
  internal: <silent> or <inner>...</inner>
  external: <silent> or <speak>...</speak>
Either, both, or neither branch may activate.

INTERNAL DECISION
Record only new query-relevant evidence or state changes that must persist.
Use concise, objective, information-additive summaries. Do not repeat stored
facts. If no useful state change occurred, output <silent>.

EXTERNAL DECISION
Respond only when the query's observable condition is satisfied. The response
must be concise, query-compliant, and supported by evidence available now.
Only <speak> content is visible to the user. If no response is warranted,
output <silent>.

OUTPUT
Emit exactly one internal decision and one external decision. Do not nest tags
or place text outside them.

QUERY-TYPE STRATEGIES AND EXAMPLES
[...]
\end{promptbox}

\begin{promptbox}{A2. Proactive-query construction and filtering (abridged)}
TASK
Generate 1--5 natural proactive requests for the observed stream, then retain
only requests that satisfy every feasibility criterion.

GENERATION CRITERIA
1. The answer depends on future visual evidence.
2. A short local clip cannot resolve the request.
3. Correct behavior requires intermediate state updates.
4. The response has a precise, observable trigger before stream termination.
5. The request is visually grounded and non-speculative.
6. The query is one sentence, normally 8--18 words, and never exceeds 20.
Do not refer to a video, clip, recording, timeline, absolute time, or ending.

FILTER IF ANY CONDITION HOLDS
requires_terminal_condition | requires_audio | unanswerable |
requires_external_knowledge

OUTPUT
Return valid JSON only. For each retained request include:
query, interaction_type, reasoning_pattern, why_proactive,
why_think_is_needed, think_steps, what_to_track, response_steps.
For filtering, return KEEP/FILTER, category, evidence-based reason, and
plausible answer times for KEEP items.

TAXONOMY, SCHEMA DETAILS, AND FEW-SHOT EXAMPLES
[...]
\end{promptbox}

\begin{promptbox}{A3. Causal dual-stream trajectory synthesis (abridged)}
TASK
Observe the scene causally. Track only what matters for the query and respond
at the earliest valid moment. A trajectory may contain one or many responses.

HIDDEN-STATE RULES
1. Include at least one state update when one is needed.
2. Update on query-relevant events, not at fixed intervals.
3. Store the minimal sufficient state: counts, entity status, or milestones.
4. Do not write process narration such as "continue monitoring".
5. Within this synthesis route, a hidden state cannot share a timestamp with,
   or preview, the next response.
6. Use an observer's perspective. Never refer to a video or recording.

RESPONSE RULES
Respond when the condition first becomes supported. Do not respond early or
late. Use only evidence observed by that time. Keep the content concise and
omit clock references from the response text.

OUTPUT
Produce only a chronological sequence:
{"think": "...", "time": "MM:SS"}
{"response": "...", "time": "MM:SS"}

FEW-SHOT TRAJECTORIES
[...]
\end{promptbox}

\begin{promptbox}{A4. Response judging and full-trajectory revision (abridged)}
EVALUATION
Given the video, query, and timestamped responses:
1. Judge each response for factual correctness and trigger timing.
2. Mark content-free repetitions or non-events for removal.
3. Find every query-relevant response opportunity that was missed.
A single-response query permits only its target event. A multi-response query
requires one response for each distinct target occurrence.

OUTPUT
Return one JSON object:
{
  "responses": [
    {"time": "...", "correct": true|false, "reason": "...",
     "proposed_time": "MM:SS|SKIP", "proposed_content": "..."}
  ],
  "missed": [{"around": "MM:SS", "reason": "..."}],
  "keep": true|false,
  "summary": "..."
}
Include proposed fields only when a correction is required.

REVISION
Re-observe the stream and regenerate the entire trajectory from scratch.
[FIX T] corrects content or timing. [REMOVE T] deletes a redundant response.
[ADD ~T] inserts a missed response. Do not patch isolated entries.

DETAILED CRITERIA AND EXAMPLES
[...]
\end{promptbox}

\begin{promptbox}{A5. Inner-state effectiveness test (abridged)}
For each target response, construct two reconstruction contexts.

A. VISUAL REFERENCE
Input: query + prior user-visible responses + frames observed since the
previous response.
Excluded: all hidden-state entries.

B. INNER-STATE CONTEXT
Input: query + prior hidden-state and response entries + only the most recent
visual window.
Excluded: earlier visual evidence.

RECONSTRUCTION
Produce one short response supported at the end of the supplied context.
If the response is not yet warranted, output exactly NO_RESPONSE_YET.

SEMANTIC JUDGE
Compare prediction and reference under the query. Return:
{"correct": true|false, "reason": "<one short sentence>"}
Wording and harmless detail may differ. A missing key event, contradiction,
different subject or action, refusal, empty output, or NO_RESPONSE_YET is
incorrect when the reference is substantive.

RETENTION RULE
Keep a trajectory only if the inner-state context reconstructs every applicable
target. The visual-reference result records whether visual-history removal
preserves capability or exposes information supplied by the hidden state.
\end{promptbox}

\clearpage
\section{Case Studies}
\label{app:case_studies}

\begin{figure}[H]
    \centering
    \providecommand{\casepanelwidth}{\linewidth}
    \captionsetup[subfigure]{skip=4pt}
    \begin{subfigure}{\casepanelwidth}
        \centering
        \includegraphics[width=\linewidth]{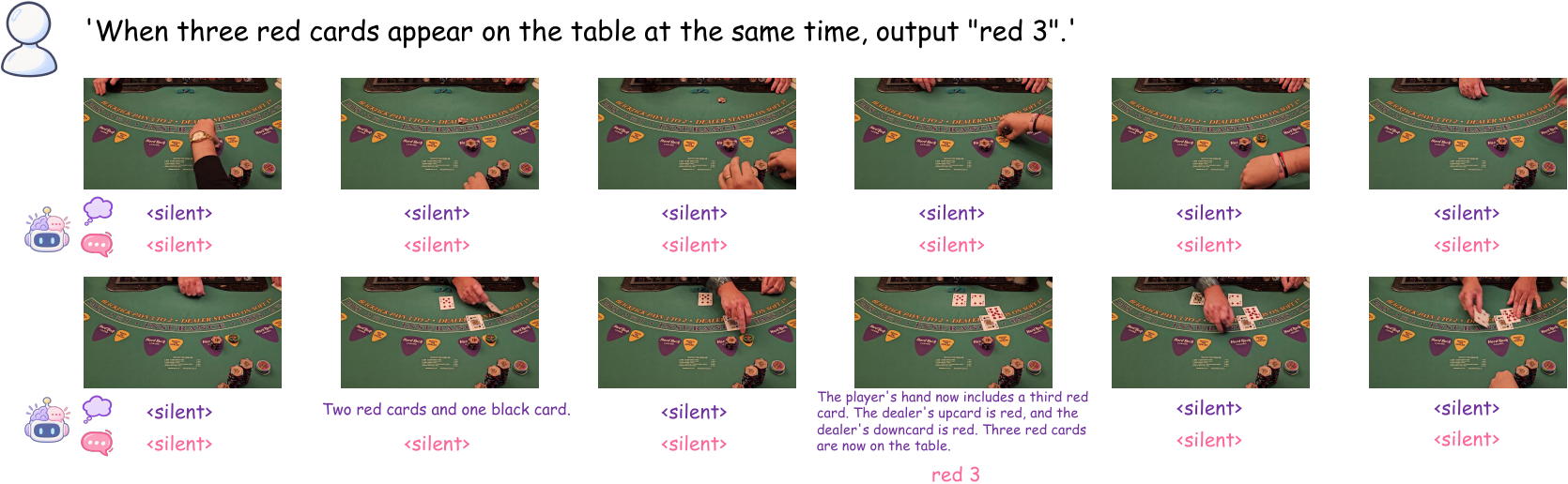}
        \caption{StreamingBench-PO.}
        \label{fig:case_sbpo}
    \end{subfigure}

    \begin{subfigure}{\casepanelwidth}
        \centering
        \includegraphics[width=\linewidth]{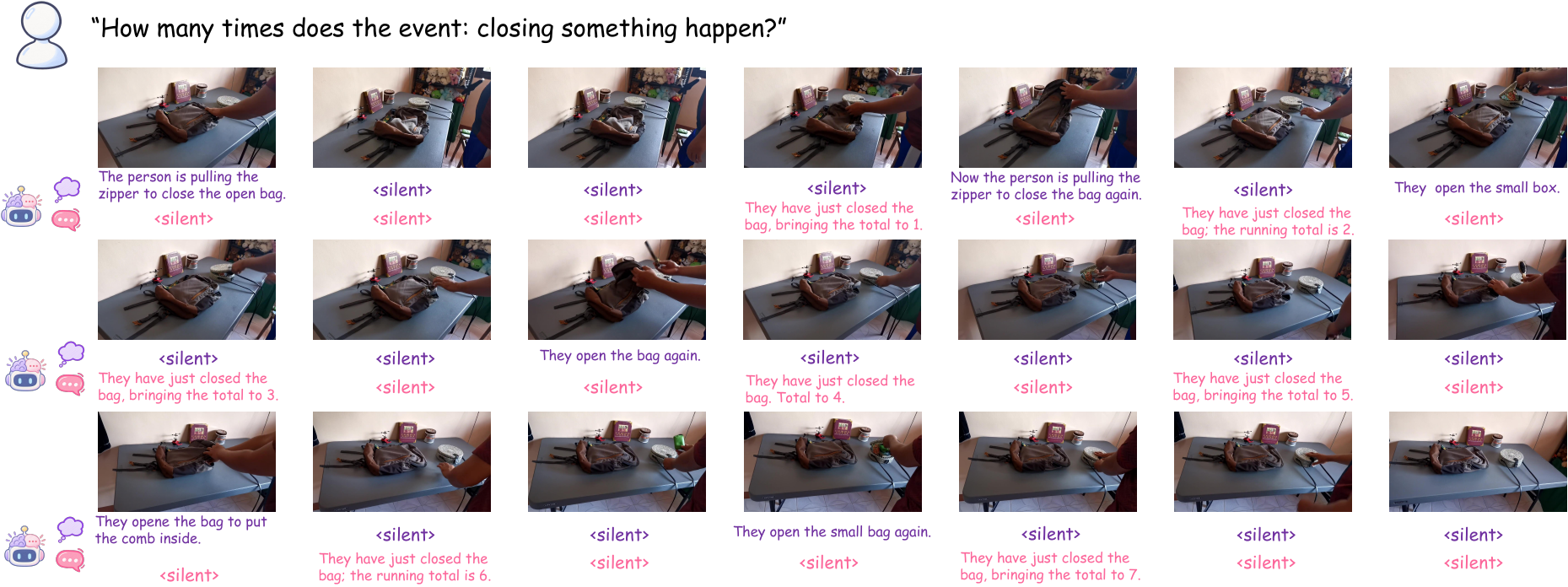}
        \caption{OVO-FAR repetition counting.}
        \label{fig:case_rec}
    \end{subfigure}

    \begin{subfigure}{\casepanelwidth}
        \centering
        \includegraphics[width=\linewidth]{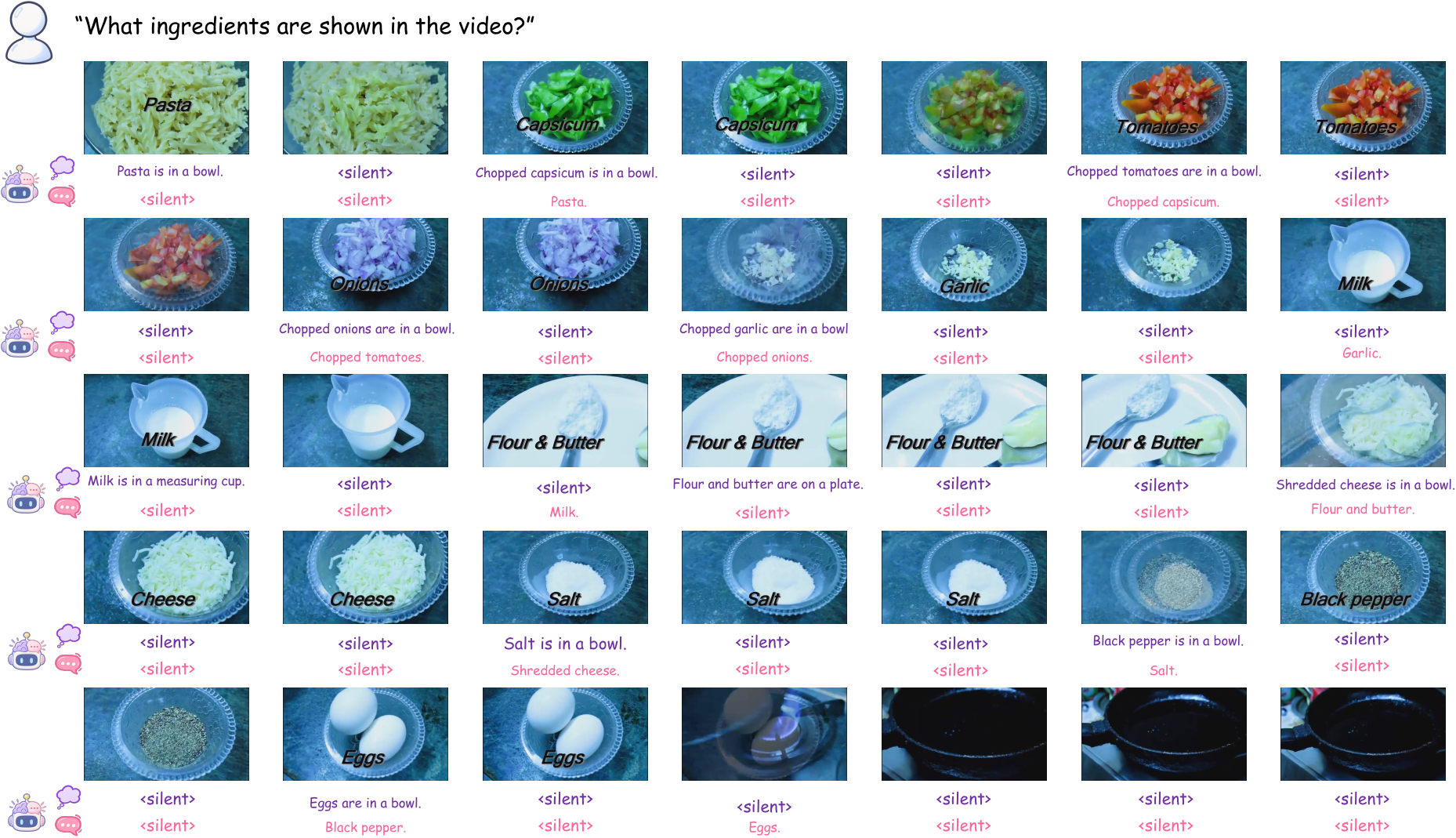}
        \caption{ProactiveVideoQA.}
        \label{fig:case_pvqa}
    \end{subfigure}
    \caption{\textbf{Case studies.} \innername{} updates are purple and \responsename{} outputs are pink.}
    \label{fig:case_studies}
\end{figure}

Figure~\ref{fig:case_studies} shows how \ours{} coordinates its two streams. \innername{} records relevant observations while \responsename{} provides the requested outputs. The branches activate together or separately with periods of silence. These behaviors illustrate how independent activation coordinates ongoing evidence tracking with selective user-facing responses.

\end{document}